\documentclass[runningheads]{llncs}

\usepackage[mobile, year=2024]{eccv}

\usepackage{eccvabbrv}

\usepackage{graphicx}
\usepackage{booktabs}

\usepackage[accsupp]{axessibility}  

\usepackage[pagebackref,breaklinks,colorlinks,citecolor=eccvblue]{hyperref}

\usepackage{orcidlink}

\usepackage{multirow}
\usepackage{bbding}
\usepackage{stfloats}

\begin{document}

\title{Toward Multi-class Anomaly Detection: Exploring Class-aware Unified Model against Inter-class Interference} 


\titlerunning{Exploring Class-aware Unified Model}

\author{Xi Jiang\inst{1} \and
Ying Chen\inst{2} \and
Qiang Nie\inst{2} \and
Jianlin Liu\inst{2} \and
Yong Liu\inst{2} \and
Chengjie Wang\inst{2} \and
Feng Zheng\inst{1}}

\authorrunning{J. Xi et al.}

\institute{Southern University of Science and Technology (SUSTech), China 
\email{jiangx2020@mail.sustech.edu.cn, zhengf@sustech.edu.cn}\\\and
Youtu Lab, Tencent, China\\
\email{\{mumuychen,stephennie,jenningsliu,choasliu,jasoncjwang\}@tencent.com}}

\maketitle

\begin{abstract}
In the context of high usability in single-class anomaly detection models, recent academic research has become concerned about the more complex multi-class anomaly detection. 
Although several papers have designed unified models for this task, they often overlook the utility of class labels, a potent tool for mitigating inter-class interference. 
To address this issue, we introduce a \textit{Multi-class Implicit Neural representation Transformer for unified Anomaly Detection (MINT-AD)}, which leverages the fine-grained category information in the training stage. By learning the multi-class distributions, the model generates class-aware query embeddings for the transformer decoder, mitigating inter-class interference within the reconstruction model. Utilizing such an implicit neural representation network, MINT-AD can project category and position information into a feature embedding space, further supervised by classification and prior probability loss functions. Experimental results on multiple datasets demonstrate that MINT-AD outperforms existing unified training models. 
\end{abstract} 

    
\section{Introduction}
\label{sec:intro}
Recently, anomaly detection (AD) and localization have been widely applied in the industrial quality inspection~\cite{cohen2020sub, yi2020patch,liu2023deep}, medical diagnosis~\cite{ding2022catching, salehi2021multiresolution}, and other fields. These applications aim to identify abnormal images and locate abnormal pixel regions. However, in real-world scenarios, only a small portion of data is abnormal, and the abnormal modality varies unpredictably, which hinders large-scale data collection and annotation. Therefore, most researches in this area are unsupervised~\cite{chen2022deep, qiu2022latent, li2022towards, jiang2022softpatch}, or self-supervised~\cite{li2021cutpaste, yao2022explicit} with artificial abnormal or generation techniques. 

Among current State-Of-The-Art (SOTA) methods~\cite{liu2023simplenet, roth2021towards}, training a separate model for each class has shown high performance in current benchmarks. However, in practical applications such as industrial quality inspection, factories may prefer a single model to detect anomalies in multiple products or multiple views of a single object. Loading all models simultaneously into memory or frequently loading and unloading models during the AD process is not feasible due to the significant memory and time consumption. 
Therefore, a more practical approach is to train a single AD model for multiple object categories. However, training different categories together presents significant challenges due to inter-class interference, particularly in highly fragmented fields such as industrial inspection, where there are a massive number of subcategories of industrial products with differences and changes in various aspects. 

\begin{figure*}[tb]
    \centering
    \includegraphics[width=0.99\linewidth]{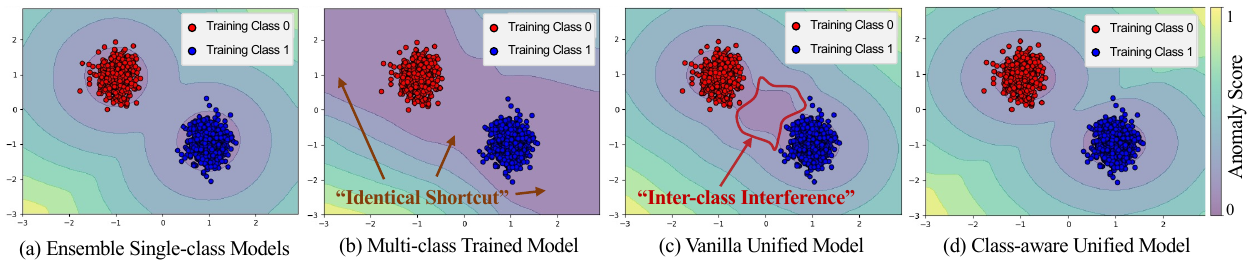}
    \caption{Decision boundary of Reconstruction Models. ``Identical shortcut'' occurs when the single-class model is trained with multi-class data. The previous unified model has significantly mitigated this issue. \textbf{``inter-class interference''} now constitutes the primary challenge facing the present unified models. }
    \label{fig:toy-dataset}
    \vspace{-0.8em}
\end{figure*}

Although existing multi-class AD models~\cite{you2022unified, Zhao_2023_CVPR, Lu2023HierarchicalVQ} have achieved the unification of anomaly detection models by addressing the ``identical shortcut" issue in the reconstruction model, they aim to capture the distribution of all classes simultaneously, allowing them to share the same boundary. 
However, this shared boundary may not exist because the feature distribution differences between different classes are already significant. 
Rather than finding a common decision boundary for multi-class scenarios, reducing the interference between each class through category information is a simple and logical method for reconstructing anomalous features. 
As shown in Fig.~\ref{fig:toy-dataset}, we created a toy dataset to show the key challenge. 
When transferring a single-class model to a multi-class task, ``identical shortcut''~\cite{liu2023diversity} can be observed. However, a vanilla unified model cannot also obtain compact boundaries because the model has generalized its reconstructive capabilities to inter-class regions, which is referred to as ``inter-class interference''. As a result, developing a unified anomaly detection model that can handle multiple object categories is a crucial but challenging task that requires innovative solutions. 


To mitigate the aforementioned issues, in this paper, we propose a novel MINT-AD method that uses an Implicit Neural Representation (INR) network to output the class-aware query of the reconstruction network. The class-aware query contains both the category information and the distribution of normal features of each category by two means: 
Classification loss function to supervise the input of the perceptron to output a fine-grained category prompt;  
Prior loss function to regularize feature adapter's outputs to Gaussian distributions and supervise the implicit neural representation network to predict the parameters of distributions. 
This paper applies the multi-class anomaly detection setting to larger datasets compared to previous research. Experimental results show that MINT-AD outperforms existing unified training models and consistently performs with the best single-class training model in most classes.

%
\noindent We summarize our main contributions as follows:
\begin{itemize}
\item We declare that ``inter-class interference'' is the crucial problem of the unified AD model, and class information can help to work it out. 

\item We propose using fine-grained features from the classifier as prompts to provide more information on subclass distribution. Then, we introduce INR as a query generator to utilize category information efficiently. 


\item Experimental results of both standard multi-class AD datasets and larger unified datasets demonstrate the effectiveness of the proposed method. 

\end{itemize}
\section{Related Work}


\subsection{Multi-class Unsupervised Anomaly Detection}
In contemporary research, unsupervised AD refers to training a model using a set of in-distribution data, which is then applied to discern between In-Distribution (ID) and Out-Of-Distribution (OOD) data originating from the same source~\cite{yang2022openood}. Ordinarily, such as a quality inspection algorithm employed on a production line~\cite{liu2024deep, jiang2022survey}, an anomaly detection model only processes a singular source. Two mainstream genres are embedding-based~\cite{defard2021padim, roth2021towards} and reconstruction-based~\cite{bergmann2018improving, sabokrou2018adversarially, liu2020towards, chen2022utrad, you2022unified} methods. 


Recent research~\cite{you2022unified} proposed the multi-class AD setting to utilize a single model to manage tasks across multiple data sources, such as employing one algorithm to inspect several production lines. This approach can conserve memory-consuming as the number of classes increases. While embedding-based methods have achieved SOTA performance in the separated setting, their memory consumption is prone to increase with the number of categories. Meanwhile, reconstruction-based methods tend to get into the "identical shortcut" quandary. UniAD~\cite{you2022unified} addressed the shortcut problem by modifying the transformer network structure and proposing a hierarchical query decoder, a neighbor masking attention module, and a feature jittering strategy. 
Following this, OmniAL~\cite{Zhao_2023_CVPR} trains the unified model with proposed panel-guided synthetic anomaly data rather than directly using normal data and HVQ-Trans~\cite{Lu2023HierarchicalVQ} utilizes a hierarchical codebook mechanism to avoid shortcuts. The diffusion model also shows good ability in this setting~\cite{Lu2023RemovingAA}. 

However, while those methods enhanced the network structure for complex distributions, they didn't consider the uniqueness of multi-class tasks in terms of categories. In practical applications, acquiring labels for the data sources is straightforward. For instance, in a factory, various products are distributed across distinct production lines, and the labels can be documented during the data collection. It is intuitively sensible to enhance the existing multi-class AD model by utilizing the information provided by labels.


\subsection{Class-aware Networks}
Networks can utilize category information to enhance performance in many aspects. Most methods utilize category information associated with the supervised task. For instance, CATN~\cite{dong2022category} is a transformer-based HOI model that uses category text information to initialize the object query. 
MCTformer~\cite{xu2022multi} proposes using multiple class queries within the transformer to capture class-specific attention, resulting in more discriminative object localization. 
CQL~\cite{xie2023category} proposes a query strongly associated with interaction categories. 

In the OOD field, NegLabel~\cite{jiang2023negative} employs many negative labels to augment OOD detection, which gets highly discriminative embeddings through the pre-trained text encoder in VLMs, e.g., CLIP.
MaxQuery~\cite{yuan2023devil} proposes to use cluster centers of object queries in mask transformers. It can be applied to medical OOD tasks since the model has already recognized and clustered a limited number of object categories.

Drawing inspiration from the aforementioned approach, we have devised three straightforward class prompting methods rooted in the Transformer architecture, which achieve limited improvement. Upon these, by harnessing finer-grained classification features and INR, we have arrived at a more efficient structure. 


\subsection{Implicit Neural Representation}

To fulfill the task of anomaly localization, our network necessitates the reconstruction of the image's features. Applying low-dimensional label information to the high-dimensional reconstruction network is far from straightforward. 
To further enhance the performance of multi-class anomaly detection, we propose using INR~\cite{park2019deepsdf, VincentSitzmann2020ImplicitNR}. 
INR is a technique that represents object shape and appearance through function approximation without explicit grid structures. This approach has been successfully applied in various fields, including computer vision, graphics, and robotics, but is novel for AD. Notably, INR has several merits that make it a robust choice for our purposes. For example, LoE~\cite{hao2022implicit} mentions that the coordinate-based network can efficiently model a distribution of high-frequency large-scale signals. Direct-PoseNet~\cite{chen2021direct} mentions that INR can easily cope with additional unlabeled data without external supervision. The salient point is that INR with gated architectures can also depict the distributions of multiple instances~\cite{mehta2021modulated}, which is necessary for multi-class AD.

\begin{figure*}[ht]
    \vspace{-0.8em}
    \centering
    \includegraphics[width=\linewidth]{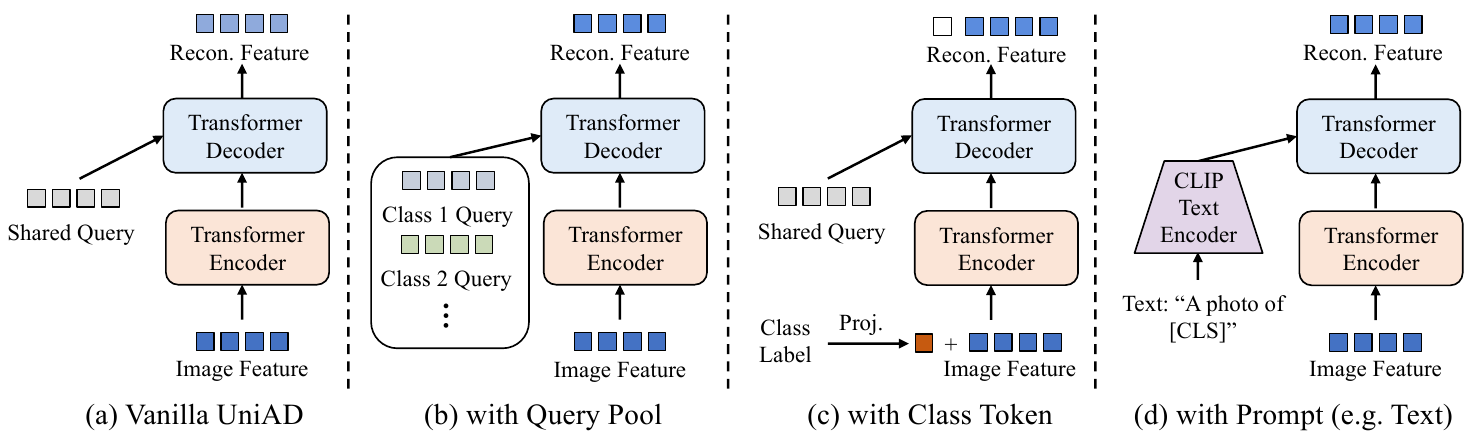}
    \caption{\textbf{Baseline and three class-aware improvements:} (a) The vanilla UniAD~\cite{you2022unified} has a shared query for all categories in each layer of the decoder. (b) One simple way is using different queries for different classes. (c) Concatenating the class token before the image feature is intuitive. (d) Using a network to map the prompt into a query can also incorporate category information. }
    \label{fig:class_embed}
    \vspace{-0.8em}
    
\end{figure*}

\section{Methods}


\subsection{Naive Class-aware Improvement}
As shown in Figure~\ref{fig:class_embed}, traditional unified networks cannot directly receive and utilize class information, but we can achieve class prompting by modifying the network structure. 
In vanilla transformer decoder of UniAD~\cite{you2022unified}, there is a learnable query embedding in each layer to prevent falling into the ``identical shortcut''. However, sharing the query embedding across different classes may exacerbate ``inter-class interference''. One simple improve approach is to establish a query pool, where different classes use different queries during the reconstruction process to address the interference caused by shared queries. The experiment substantiates the efficacy of such minor modifications. 

However, query pool will expands along with the number of classes increasing. Another direct improve approach is to utilize class tokens, which implicitly delegates the task of distinguishing between classes to the self-attention mechanism. We project category labels onto a one-dimensional token, which is then concatenated with the input image features. Subsequently, in the reconstructed features, this class token is excised. But optimizing this token projection through reconstruction loss proves to be rather challenging. 

We are curious to ascertain whether a pre-trained network exists capable of accomplishing this mapping task.
Our third method construct class-related text and use existing text encoders (from CLIP~\cite{radford2021learning}) to extract text feature, which subsequently form the query embedding through duplication. 

Although exemplify mainstream, the above three methods share common shortcomings: Firstly, they require category labels for objects during both the training and inference phases. Secondly, they neglect to consider the associative properties of image coordinates. 
Existing work has shown that images~\cite{chen2021crossvit} and text~\cite{rombach2022high} can serve as such prompts. So, we further explore the type of prompt and the structure for mapping.

\subsection{Overview of Model}
\label{adapter}
\begin{figure*}[htb]
    \centering
    \includegraphics[width=\linewidth]{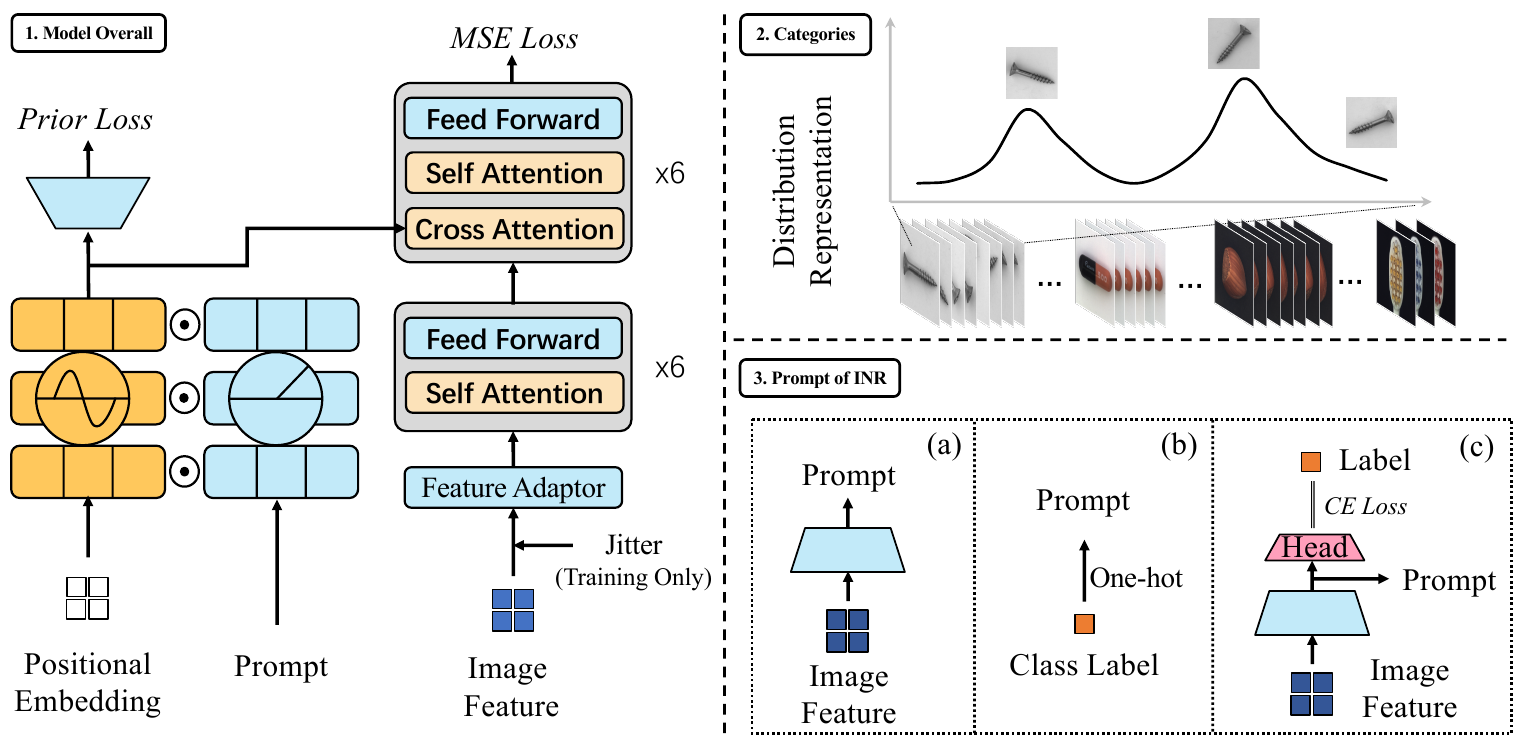}
    \caption{\textbf{Architecture of MINT-AD network}. 1. A dual-path INR network with different activation functions is introduced to map position encoding and class-aware prompt to the image feature dimension and assist the reconstruction transformer through cross-attention as queries. 2. We are trying to find a fine-grained prompt that can map the distribution of subcategories. 3. Compared with getting prompts from the (a) image or (b) label, (c) subcategory features from the classification network from the image to label is a better prompt. 
    }
    \vspace{-0.8em}
    
    \label{fig:MINT}
\end{figure*}

We introduce MINT-AD, a novel unified anomaly detection and reconstruction model based on the Transformer framework, which leverages class-specific object-aware reconstruction. The overall architecture of MINT-AD is illustrated in Fig.~\ref{fig:MINT}.1. 
The main reconstruction network is a transformer with a 6-layer encoder and decoder. We use INR to output the class-aware query and a distribution decoder to predict the distribution parameters of the adapted image feature. Using INR in anomaly detection tasks is innovative, and we believe it is a better way to incorporate class information. 

We adopted the configuration of UniAD~\cite{you2022unified}, utilizing an identical feature extractor, a same-size transformer reconstruction, and the same feature jittering. However, we did not employ the Neighbor mask and Layer-wise query as we found that feature jittering adequately addresses the concern of identity shortcut. Building upon these foundations, we introduced a query generation network to replace original query embedding. 
The feature extractor is an EfficientNet-b4~\cite{tan2019efficientnet} pre-trained with ImageNet-1K, which is frozen during training. 
Following the research that points out pre-trained feature extractors may not be optimal~\cite{liu2023simplenet,zou2022spot,Reiss2021PANDAAP}, we employ a feature adaptor (a single fully connected layer by default, omitted in the figure) to transform the features into a specific domain. We denote the original image feature as $\boldsymbol{f'}$ and the adapted feature as $\boldsymbol{f}$, while the single patch feature among them is represented by $\phi '$ and $\phi$, respectively. The detailed parameters of the model architecture can be referenced in the appendix.

\subsection{Class-aware Implicit Neural Representation}
\label{INR}
\begin{figure*}[tbh]
    \vspace{-0.8em}

    \centering
    \includegraphics[width=0.96\linewidth]{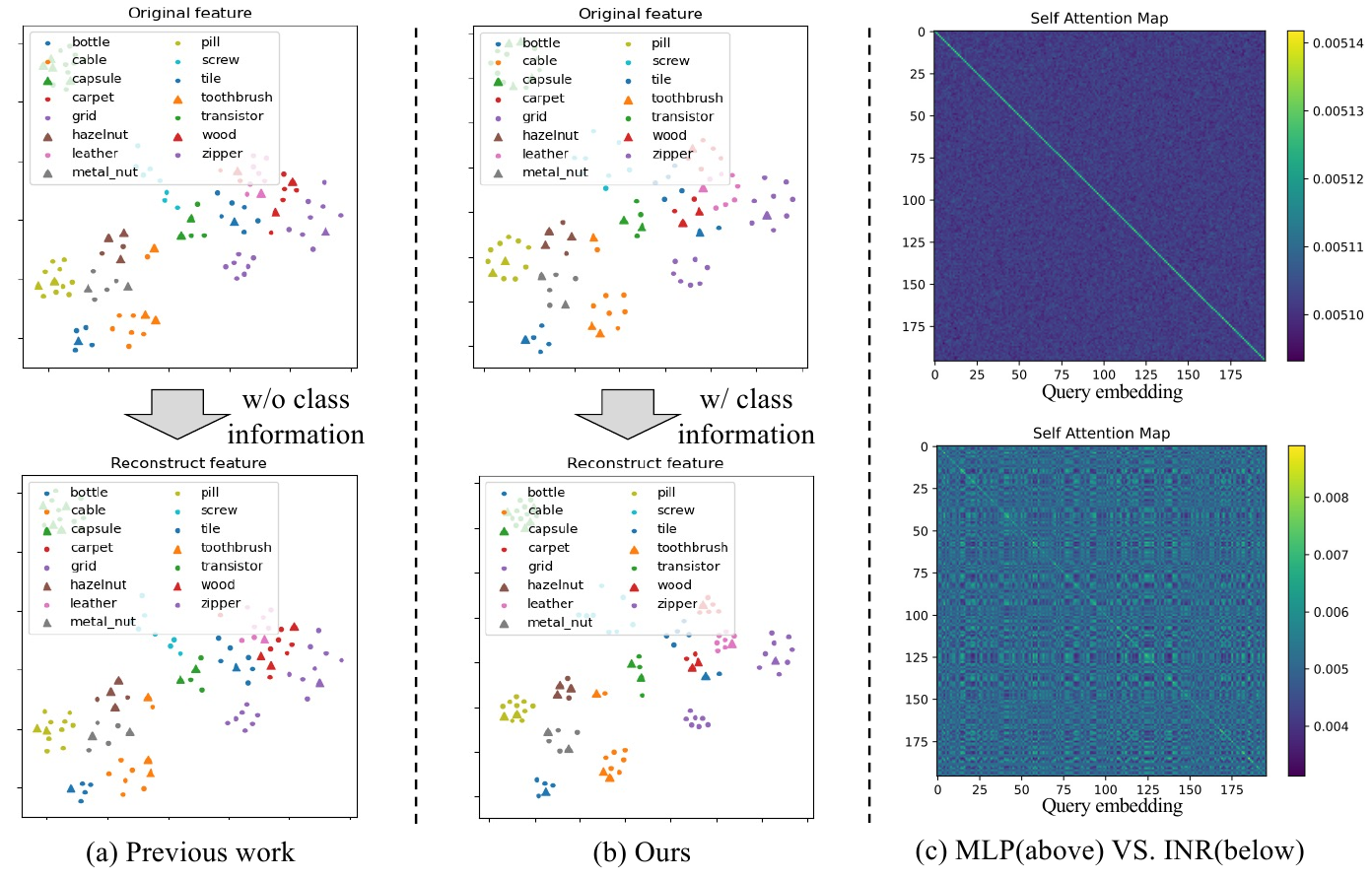}
    \caption{(a$\&$b) T-SNE visualization of image features: triangles for normal samples, circles for anomalies. Unlike prior work, our method leverages category data, making the distinctions between different classes more pronounced post-reconstruction, as the outlier samples align closer to their respective normal class counterparts. (c) The self-attention map of queries from MLP and INR. The coordinate-based network architecture of INR can model more structural information. 
    }
    \vspace{-0.8em}
    
    \label{fig:visualization}
\end{figure*}

To ensure that the reconstruction model can distinguish between multiple categories of data and that the reconstruction of each category is not affected by other categories, this paper proposes using different query values for different categories to make the reconstruction process robust to multiple categories. In the context of the feature reconstruction visualization illustrated in Fig.~\ref{fig:visualization}(a$\&$b), the previous approach fails to distinguish features among multiple categories. During training, to completely reconstruct normal features from different classes, the anomalous feature space between categories is also faithfully reconstructed, thereby impeding anomaly detection capabilities. Our method, upon incorporating category features, steers anomalous features of each class towards normalcy post-reconstruction, significantly reducing inter-class overlap. 

\noindent\textbf{Classfication Representation}. Since we cannot guarantee that each data sample has a category label, a classifier is used to predict the category of each sample. The classifier uses adapted feature $\boldsymbol{f}$ as input. After passing through the category encoder $E_{class}$ and the classification head, it obtains the predicted value $\hat{y}=head(E_{class}(\boldsymbol{f}))\in \mathbb{R}^{N}$, which is then used to calculate the cross-entropy loss with the one-hot encoded category label $y \in \mathbb{R}^{N}$ as shown in Equation~\ref{eq:celoss}:
\begin{equation}\label{eq:celoss}
    \mathcal{L}_{CE}=-\frac{1}{N}\sum_{c=0}^{N-1} y_{c} \log \hat{y}_{c}. 
\end{equation}
We found that the granularity of category labels in industrial anomaly detection is not fine enough. For example, in the MVTec AD dataset, the ``toothbrush" category includes various colors, and the ``screw" category has multiple placement postures. Therefore, we propose to use finer-grained classification features as prompts for the unified model. 
As shown in Fig.~\ref{fig:MINT}.3(c), this paper does not directly use the predicted category label or image feature but instead uses the token $t_{c} = E_{class}(\boldsymbol{f}) \in \mathbb{R}^{N'}$ before the last classification head as the representation of the subcategory. 

\noindent\textbf{Implicit Neural Representation}. 
The key issue addressed in this section is mapping from low-dimensional logistic regression to high-dimensional queries. 
We propose that the simple network, such as Multi-Layer Perceptron (MLP), is inefficient, which requires a massive number of parameters to accomplish this high-dimensional mapping. Implicit neural representation~\cite{mildenhall2021nerf}, on the other hand, is well-suited for solving this high-dimensional mapping problem from discrete values to continuous values. INR is a continuous function mapping position encoding to values, and to enable it to map different categories, we use the concept of multiple instances of Neural Radiance Fields and employ a dual-path structure. Like the position encoding in Transformer, the position encoding uses a learnable embedding. 

Similar to~\cite{mehta2021modulated}, the dual-path structures use different activation functions to achieve different functions. The MLP that takes the position encoding as input is called the synthesis network, which implements the mapping from coordinates to signal values and uses sine functions as activation functions, like SIREN~\cite{VincentSitzmann2020ImplicitNR}. The MLP that takes the category encoding as input is called the modulator, which uses a ReLU-based function activation. The modulator is the key to category generalization, and it outputs the amplitude, phase, and frequency parameters of the periodic activation in the modulated synthesis network at each layer. The formulas are expressed in the appendix.

INR can model discrete construction by mapping position encoding to the query embedding. This implies a superior capacity to comprehend structural information among distinct categories of image features. We elucidate this aspect in Fig.~\ref{fig:visualization}(c), which visualizes the self-attention map of the queries outputted by the network. Only the diagonal elements are prominent in the self-attention map corresponding to the MLP. This observation indicates that while the MLP accomplishes a mapping to high-dimensional image feature space, it fails to capture inter-feature relationships. Given that image feature maps correspond to various regions within an image, although the optimal level of self-correlation remains unclear, the self-attention map associated with the INR framework at least reveals that its query embedding output encapsulates some structural characteristics. 

Moreover, the parameters of MLP count surpass INR by over 500 times due to the larger output dimension, with a slightly inferior performance.  
Implicit neural networks can effectively map category information to the image feature dimension. However, we expect the corresponding feature space to be continuous. Additionally, class token $t_c$ should correspond to the entire subclass of samples rather than individual instances. Therefore, we further design prior distribution representation and prior loss functions to supervise the mapping process. 

\subsection{Prior Distribution Representation}

Since each subclass represents a distribution of normal images, we introduce an additional decoder to map the features output by INR into parameters for Gaussian distributions of image features. 
Assuming that all image features within the same subclass $\{\boldsymbol{f'}_c\}$ can be projected onto the same token $t_{c}$, we can use a predefined distribution $q(\omega_c)$ to estimate the statistics of features, where the parameters $\omega_c$ of this distribution can be derived from class token mappings. 
With $w_c$ getting from neural networks, The parameters of whole distribution networks $\theta_p$ are estimated with maximum likelihood estimation:
\begin{equation}
{\theta_p}^*=\arg \max _{\theta_p} \frac{1}{|\mathcal{D}|} \sum_{\boldsymbol{f} \in \mathcal{D}} \log \mathcal{P}_{\theta_p}\left( \boldsymbol{f'}_c \mid q(\omega_c)\right).
\end{equation}
However, the distribution functions for different categories may vary. To standardize the distribution of image features, we propose using a network to adapt the image features. We reuse the adapter mentioned in Section \ref{adapter} to transform the estimation of original image features into an estimation of adapted feature distribution. 
The log-likelihood can be rewritten as
\begin{equation}
\begin{split}
\log \mathcal{P}_{\{\theta_p, \theta_a\}}( \boldsymbol{f}_c \mid & q(\omega_c))  = \mathcal{L}_{E L B O}(q, \theta_a) \\
&+ KL[q(\omega_c \mid t_{c},\theta_p) \| p_{\theta_a}(\boldsymbol{f'}_c)], 
\end{split}
\end{equation}
where $\theta_a$ is parameter of adapter, and $q(\omega_c)$ try to approximate the adapted feature distribution $\mathcal{P}_{\theta_a}(\boldsymbol{f'}_c)$. The feature adaptor optimizes the evidence lower bound $\mathcal{L}_{E L B O}$, and distribution projection optimizes the Kullback-Leibler divergence $KL[.]$. So, these two modules can be optimized by a single prior loss function. 
The preset distribution $q$ for the entire image features can be set as the joint distribution of distributions for each position. 
Ignoring the conditional dependence, we set the distribution in each position as an independent normal distribution. 
The decoder that maps the distribution only needs to output the mean $\mu$ and variance $\sigma$ for each position. The prior loss can be defined as 
\begin{equation}\label{eq:priorloss}
    \mathcal{L}_{Prior}=\sum_{\phi \in \boldsymbol{f}} -\log \mathcal{P}(\phi | \mathcal{N}(\mu,\sigma)).
\end{equation}

\subsection{Loss Function}
The output mapped by INR from fine-grained category features will serve as the query for the reconstruction network and interact with the image features in the cross-attention of each decoder layer. With the introduction of category information, the reconstruction network can reduce interference between categories. We maintain the mean squared error (MSE) loss function as the reconstruction supervisor: 
\begin{equation}\label{eq:mseloss}
\mathcal{L}_{MSE}=\frac{1}{H \times W} |\boldsymbol{f'}- \hat{\boldsymbol{f}}|^2_2,
\end{equation}
where $\hat{\boldsymbol{f}}$ is the reconstructed feature. 

The overall loss function is shown in Equation~\ref{eq:total_loss}, where $\lambda_a$ and $\lambda_b$ are weighting coefficients, which are set to 1 and 0.1, respectively, to optimize the classification and reconstruction networks first, as classification accuracy is a prerequisite for isolating reconstruction of each category, and the Transformer reconstruction network is more difficult to optimize.
\begin{equation}\label{eq:total_loss}
    \mathcal{L}_{total}= \mathcal{L}_{MSE} + \lambda_a \mathcal{L}_{CE} + \lambda_b \mathcal{L}_{Prior} .
\end{equation}


\subsection{Inference and Anomaly Score Function}
During the testing phase, except for not adding perturbations, all other parts of the network remain unchanged. Since the prior loss is a relatively fuzzy mapping, the reconstruction error is directly used as the anomaly score. For anomaly localization, the anomaly image is composed of the reconstruction distance of the feature map and is linearly interpolated to the size of the image input. The anomaly score of each patch feature before interpolation is:
\begin{equation}
s_{hw} = |\phi '- \hat{\phi}|^2_2 .
\end{equation}
For anomaly detection, the image-level anomaly score is the maximum value of the pixel-level anomaly score, which is generated by linearly interpolating the anomaly map of the feature dimension.
\section{Experiments}

\subsection{Datasets and Metrics}
\textbf{Datasets}. We expand on previous research by conducting experiments on a wider range of sensory datasets, including the commonly used MVTec AD~\cite{bergmann2019mvtec} and the newly proposed VisA~\cite{zou2022spot} and BTAD~\cite{mishra2021vt}. The separated setting and unified setting are following existing papers~\cite{you2022unified,Zhao_2023_CVPR,Lu2023HierarchicalVQ}. By testing these datasets, we demonstrate the effectiveness and robustness of the proposed method in detecting anomalies in various industrial settings.
Besides, we use CIFAR-10 to test the performance of semantic AD~\cite{lee2023selformaly}.



\noindent\textbf{Evaluation Metrics}.
We calculate each category's image-level AUROC (I-AUROC) and report the average of 3 runs. In addition, to better demonstrate the anomaly localization ability of each method~\cite{zou2022spot, xie2023iad}, we use pixel-level AUPR (P-AUPR) to measure the localization performance, which focuses more on the abnormal regions. 
AUPR is also referred to as AP and primarily focuses on the performance of positive classes. 
Several studies, such as spot-diff~\cite{zou2022spot}, have highlighted the influence of different metrics and why AUPR is considered a more informative choice in pixel-wise anomaly detection.

\noindent\textbf{Implementation Details.}
This section describes the configuration implementation details of the experiments in this paper. 
The feature maps are extracted from the first to the fourth stages of EfficientNet-b4 (pre-trained in ImageNet). The extracted image feature map size is set to $14\times14$. 
We use the AdamW optimizer with weight decay of $1\times10^{-4}$. All other parts are trained from random initialization except for the pre-trained feature extractor. The model is trained for 1000 full epochs with a batch size of 32 on 4 GPUs (NVIDIA Tesla V100 32GB). 
The initial learning rate is set to $1\times10^{-4}$ and is decreased by 0.1 after 800 full epochs. 

\begin{table*}[ht]
    \vspace{-0.8em}

\centering
\setlength\tabcolsep{3pt}
\caption{Performance comparison of different methods in anomaly detection with MVTec AD. (Metrics: I-AUROC of Unified/\textcolor{lightgray}{Separate})}
\resizebox{\linewidth}{!}{
\begin{tabular}{ll|ccccccc}
\toprule\toprule
\multicolumn{2}{c|}{Categories} & PaDiM~\cite{defard2021padim} & CutPaste~\cite{li2021cutpaste} & DRAEM~\cite{7} & SimpleNet~\cite{liu2023simplenet} & UniAD~\cite{you2022unified} & OmniAL~\cite{Zhao_2023_CVPR} & MINT-AD(ours) \\ \midrule
\multirow{10}{*}{\rotatebox{90}{Objects}} 
&Bottle & 0.979/\textcolor{lightgray}{0.992} & 0.679/\textcolor{lightgray}{0.982} & 0.946/\textcolor{lightgray}{0.991} & 0.908/\textcolor{lightgray}{1.000} & 0.997/\textcolor{lightgray}{1.000} & \textbf{1.000}/\textcolor{lightgray}{0.994} & \textbf{1.000}/\textcolor{lightgray}{0.992} \\
&Cable & 0.709/\textcolor{lightgray}{0.971} & 0.692/\textcolor{lightgray}{0.812} & 0.618/\textcolor{lightgray}{0.947} & 0.561/\textcolor{lightgray}{0.999} & 0.952/\textcolor{lightgray}{0.976} & \textbf{0.982}/\textcolor{lightgray}{0.976} & 0.935/\textcolor{lightgray}{0.986} \\
&Capsule & 0.734/\textcolor{lightgray}{0.875} & 0.630/\textcolor{lightgray}{0.982} & 0.702/\textcolor{lightgray}{0.985} & 0.777/\textcolor{lightgray}{0.974} & 0.869/\textcolor{lightgray}{0.853} & 0.952/\textcolor{lightgray}{0.924} & \textbf{0.955}/\textcolor{lightgray}{0.810} \\
&Hazelnut & 0.855/\textcolor{lightgray}{0.994} & 0.809/\textcolor{lightgray}{0.983} & 0.951/\textcolor{lightgray}{1.000} & 0.820/\textcolor{lightgray}{1.000} & 0.998/\textcolor{lightgray}{0.999} & 0.956/\textcolor{lightgray}{0.980} & \textbf{1.000}/\textcolor{lightgray}{1.000} \\
&Metal nut & 0.880/\textcolor{lightgray}{0.962} & 0.600/\textcolor{lightgray}{0.999} & 0.889/\textcolor{lightgray}{0.987} & 0.501/\textcolor{lightgray}{1.000} & \textbf{0.992}/\textcolor{lightgray}{0.990} & 0.992/\textcolor{lightgray}{0.999} & \textbf{0.993}/\textcolor{lightgray}{0.971} \\
&Pill & 0.688/\textcolor{lightgray}{0.901} & 0.714/\textcolor{lightgray}{0.949} & 0.690/\textcolor{lightgray}{0.989} & 0.595/\textcolor{lightgray}{0.990} & 0.937/\textcolor{lightgray}{0.883} & \textbf{0.972}/\textcolor{lightgray}{0.977} & 0.958/\textcolor{lightgray}{0.983} \\
&Screw & 0.569/\textcolor{lightgray}{0.975} & 0.852/\textcolor{lightgray}{0.887} & 0.933/\textcolor{lightgray}{0.939} & 0.486/\textcolor{lightgray}{0.981} & 0.875/\textcolor{lightgray}{0.919} & 0.880/\textcolor{lightgray}{0.810} & \textbf{0.983}/\textcolor{lightgray}{0.978} \\
&Toothbrush & 0.953/\textcolor{lightgray}{1.000} & 0.639/\textcolor{lightgray}{0.994} & 0.828/\textcolor{lightgray}{1.000} & 0.608/\textcolor{lightgray}{0.997} & 0.942/\textcolor{lightgray}{0.950} & \textbf{1.000}/\textcolor{lightgray}{1.000} & \textbf{1.000}/\textcolor{lightgray}{0.956} \\
&Transistor & 0.866/\textcolor{lightgray}{0.944} & 0.579/\textcolor{lightgray}{0.961} & 0.839/\textcolor{lightgray}{0.931} & 0.678/\textcolor{lightgray}{1.000} & \textbf{0.998}/\textcolor{lightgray}{1.000} & 0.938/\textcolor{lightgray}{0.938} & \textbf{0.998}/\textcolor{lightgray}{0.999} \\ \midrule
&Zipper & 0.797/\textcolor{lightgray}{0.986} & 0.935/\textcolor{lightgray}{0.999} & 0.991/\textcolor{lightgray}{1.000} & 0.837/\textcolor{lightgray}{0.999} & 0.958/\textcolor{lightgray}{0.967} & \textbf{1.000}/\textcolor{lightgray}{1.000} & 0.991/\textcolor{lightgray}{0.996} \\
\multirow{5}{*}{\rotatebox{90}{Textures}}  
&Carpet & 0.938/\textcolor{lightgray}{0.998} & 0.936/\textcolor{lightgray}{0.939} & 0.959/\textcolor{lightgray}{0.955} & 0.902/\textcolor{lightgray}{0.997} & 0.998/\textcolor{lightgray}{0.999} & 0.987/\textcolor{lightgray}{0.996} & \textbf{0.999}/\textcolor{lightgray}{1.000} \\
&Grid & 0.739/\textcolor{lightgray}{0.967} & 0.932/\textcolor{lightgray}{1.000} & 0.981/\textcolor{lightgray}{0.999} & 0.609/\textcolor{lightgray}{0.997} & 0.982/\textcolor{lightgray}{0.985} & 0.999/\textcolor{lightgray}{1.000} & \textbf{1.000}/\textcolor{lightgray}{0.999} \\
&Leather & 0.999/\textcolor{lightgray}{1.000} & 0.934/\textcolor{lightgray}{1.000} & 0.999/\textcolor{lightgray}{1.000} & 0.859/\textcolor{lightgray}{1.000} & \textbf{1.000}/\textcolor{lightgray}{1.000} & 0.990/\textcolor{lightgray}{0.976} & \textbf{1.000}/\textcolor{lightgray}{1.000} \\
&Tile & 0.933/\textcolor{lightgray}{0.981} & 0.886/\textcolor{lightgray}{0.946} & 0.983/\textcolor{lightgray}{0.996} & 0.902/\textcolor{lightgray}{0.998} & 0.993/\textcolor{lightgray}{0.990} & \textbf{0.996}/\textcolor{lightgray}{1.000} & 0.992/\textcolor{lightgray}{0.990} \\
&Wood & 0.721/\textcolor{lightgray}{0.992} & 0.804/\textcolor{lightgray}{0.991} & \textbf{0.998}/\textcolor{lightgray}{0.991} & 0.907/\textcolor{lightgray}{1.000} & 0.986/\textcolor{lightgray}{0.979} & 0.932/\textcolor{lightgray}{0.987} & 0.994/\textcolor{lightgray}{0.989} \\ \midrule
&\textbf{Mean} & 0.842/\textcolor{lightgray}{0.958} & 0.775/\textcolor{lightgray}{0.961} & 0.887/\textcolor{lightgray}{0.980} & 0.730/\textcolor{lightgray}{0.994} & 0.965/\textcolor{lightgray}{0.966} & 0.972/\textcolor{lightgray}{0.970} & \textbf{0.986}/\textcolor{lightgray}{0.977} \\ \bottomrule\bottomrule
    \end{tabular}
}
    \vspace{-0.8em}

\label{tab-mint-mvtec}
\end{table*}

\subsection{Anomaly Detection and Localization on MVTec-AD}
The results of anomaly detection on the MVTec-AD dataset are presented in Table~\ref{tab-mint-mvtec}. 
In the unified setting, although UniAD~\cite{you2022unified} and OmniAL~\cite{Zhao_2023_CVPR} are both unified reconstruction networks, MINT-AD significantly improves the performance of most categories. What's more, MINT-AD demonstrates even more significant advantages for categories that can be further subdivided, such as Toothbrush, Capsule, and Screw. This indicates that our proposed algorithm can reduce interference in both inter-class and intra-class and has better normal reconstruction ability from a classification perspective. 
While the performance of the separate methods is still inferior to that of the current best separate method, our proposed method significantly outperforms the Synthesizing-based method CutPaste~\cite{li2021cutpaste} and the Ensembling-based method PaDiM~\cite{defard2021padim}. It is worth noting that the SOTA separate method, SimpleNet, performs poorly with unified training because SimpleNet uses a discriminator model, and simply fusing multi-class information together for training would result in a significant drop in performance. In contrast, our proposed method, MINT, performed best under the unified framework. Furthermore, our approach exhibits a distinctive characteristic whereby its performance in multi-class unified training surpasses that of separate training runs. This suggests that our model has gleaned shared knowledge across different categories. Based on the qualitative results in the appendix, there is an enhancement in the robustness against noise in the background.

\begin{table*}[ht]
\centering
\caption{Anomaly detection and localization performance on VisA dataset. We further highlight the best unified-training method in bold. (Metrics: I-AUROC/P-AUPR)}
\label{tab-mint-VisA-image}
\resizebox{\textwidth}{!}{
\begin{tabular}{ll|cccc|ccc}
\toprule\toprule
\multicolumn{2}{c|}{Methods}     & \multicolumn{4}{c|}{Separate}  & \multicolumn{3}{c}{Unified}        \\\midrule
\multicolumn{2}{c|}{Categories}  & PaDiM~\cite{defard2021padim}   & CutPaste~\cite{li2021cutpaste} & PatchCore~\cite{li2022towards} &SimpleNet~\cite{liu2023simplenet} & SimpleNet~\cite{liu2023simplenet} & UniAD~\cite{you2022unified}          & MINT-AD(ours)     \\\midrule
\multirow{4}{*}{\rotatebox{90}{Single}} 
& Cashew      & 0.930/0.522     & 0.848/-     & 0.973/0.585    & 0.980/0.603 & 0.922/\textbf{0.520}&\textbf{0.928}/0.502 & 0.882/0.286 \\
& Chewing gum & 0.988/0.563     & 0.926/-     & 0.991/0.428    &0.998/0.183  &0.959/0.357 &0.996/0.604          & \textbf{0.999}/\textbf{0.624} \\
& Fryum       & 0.886/0.433     & 0.940/-     & 0.962/0.372    &0.987/0.367 &0.913/0.349 &0.898/\textbf{0.476}          & \textbf{0.954}/0.387 \\
& Pipe fryum  & 0.970/0.633     & 0.769/-     & 0.998/0.585    &0.998/0.591 &0.803/\textbf{0.628} &0.978/0.549          & \textbf{0.982}/0.471 \\
                      \midrule
                      \multirow{4}{*}{\rotatebox{90}{Multiple}}

& Candles     & 0.916/0.176     & 0.699/-     & 0.986/0.176    &0.985/0.141 &0.959/0.078 & 0.969/0.212          & \textbf{0.975}/\textbf{0.278} \\
& Capsules    & 0.707/0.182     & 0.790/-     & 0.816/0.687    &0.909/0.577 &0.731/0.442 &0.721/0.513          & \textbf{0.899}/\textbf{0.579} \\
& Macaroni1   & 0.870/0.107     & 0.934/-     & 0.975/0.076   &0.995/0.074  &0.914/0.044 & 0.918/0.088          & \textbf{0.991}/\textbf{0.240} \\
& Macaroni2   & 0.705/0.011     & 0.836/-     & 0.781/0.035    &0.823/0.049 &0.604/0.010 &0.856/0.039          & \textbf{0.901}/\textbf{0.158} \\
                      \midrule
                      \multirow{4}{*}{\rotatebox{90}{Complex}}

& PCB1        & 0.947/0.519     & 0.831/-     & 0.985/0.917    &0.996/0.849 &0.910/\textbf{0.943} &0.957/0.674          & \textbf{0.965}/0.782 \\
& PCB2        & 0.885/0.138     & 0.449/-     & 0.973/0.143   &0.995/0.131  &0.915/0.148 &0.934/0.089          & \textbf{0.964}/\textbf{0.277} \\
& PCB3        & 0.910/0.288     & 0.900/-     & 0.979/0.389    &0.992/0.442 &0.889/0.303 &0.894/0.149          & \textbf{0.980}/\textbf{0.340} \\
& PCB4        & 0.975/0.184     & 0.907/-     & 0.996/0.424    &0.995/0.378 &0.961/0.331 &\textbf{0.994}/\textbf{0.365}          & 0.988/0.304 \\
                      \midrule
& {\textbf{Mean}}         & 0.891/0.309  & 0.819/- & 0.951/0.401     &0.971/0.365 &0.873/0.338 &0.920/0.355     & \textbf{0.957}/\textbf{0.394}\\
\bottomrule\bottomrule
    \vspace{-0.8em}

\end{tabular}}
\end{table*}
\subsection{Anomaly Detection and Localization on VisA} 
In order to conduct a more comprehensive analysis, we opted for the VisA~\cite{zou2022spot} dataset, which also incorporates categories with multiple instances and complex structures, going beyond simple single-instance anomaly detection. 
Upon analyzing the results in Table~\ref{tab-mint-VisA-image}, MINT-AD achieves a larger improvement than UniAD in the unified setting in VisA compared with MVTec AD. By comparing the I-AUROC and P-AUPR metrics for each category, we found that the advantage of MINT-AD mainly stems from its more balanced performance across all categories. This implies a reduction in `` inter-class interference''. 
In the separate setting, we found that SimpleNet~\cite{liu2023simplenet} still achieved SOTA performance in I-AUROC. Despite training with multi-class, our method has just a slightly lower I-AUROC than separate SimpleNet, but a higher P-AUPR. 
Additionally, we observed that for some non-rigid objects, such as Macaroni, the P-AUPR of all methods was very low due to the shape of the object changing to some extent. 
Overall, our method demonstrated better adaptability to complex scenes and achieved the best performance in the unified methods. 

\subsection{Performance in Larger Dataset}
\label{larger-dataset}
To explore how the unified models perform with an increased number of categories, we conducted training using a larger merged dataset consisting of the MVTec, VisA, and BTAD anomaly detection datasets. The corresponding results are presented in Table~\ref{tab-mint-all}, where the superiority of MINT-AD over UniAD becomes more evident. Specifically, when faced with the pressure of unexpected multi-class scenarios, our method still experiences certain performance degradation, with a decrease of 0.018 in I-AUROC on MVTec AD and a decrease of 0.033 on VisA. This may be due to the limited ability of the reconstruction model to fit fine-grained feature details. UniAD uses an equal-sized base reconstruction model but performs worse. This demonstrates that our class-aware network has better robustness when the number of classes increases. 

\begin{table}[hbt]
\setlength\tabcolsep{3pt}
    \centering
    \caption{Performance comparison of algorithms on a unified Dataset Comprising MVTec, VisA, and BTAD where 30 categories are combined for unified training. (I-AUROC/P-AUPR)}
    \label{tab-mint-all}
    \resizebox{0.7\linewidth}{!}{
    \begin{tabular}{l|ccc|c}
    \toprule\toprule
        Methods & BTAD~\cite{mishra2021vt} & MVTec~\cite{bergmann2019mvtec} & VisA~\cite{zou2022spot} & Mean \\ \midrule
        UniAD & 0.921/\textbf{0.469}  & 0.936/0.416  & 0.906/0.336  & 0.923/0.389  \\ 
        MINT-AD & \textbf{0.954}/0.445  & \textbf{0.969}/\textbf{0.450 }  & \textbf{0.924}/\textbf{0.359}  & \textbf{0.949}/\textbf{0.413} \\ \bottomrule\bottomrule
    \end{tabular}}
    \vspace{-0.8em}
\end{table}

\begin{table}[ht]
\centering
\caption{Different class-aware methods. \Checkmark represents using labels in the training/testing stage.}
\resizebox{0.7\linewidth}{!}{
    \begin{tabular}{lcccc}\\\toprule\toprule
    \multirow{2}{*}{Methods} & \multicolumn{2}{c}{Using Label} & \multirow{2}{*}{I-AUROC} & \multirow{2}{*}{P-AUPR}\\ \cline{2-3} 
    & Training & Testing & & \\
    \midrule
    Vanilla UniAD  & - & - & 0.965 & 0.445\\
    with Query Pool & \Checkmark & \Checkmark & 0.976  & 0.470\\  
    with Class Token & \Checkmark & \Checkmark & 0.979  & 0.470\\  
    with Text Prompt  & \Checkmark & \Checkmark & 0.977  & 0.463\\   \midrule 
    \emph{Prompt Type of MINT-AD}&&& \\
    Image Prompt & - & - & 0.978 & 0.465 \\ 
    Label Prompt & \Checkmark & \Checkmark & 0.980 & 0.472 \\ 
    Feature Prompt & \Checkmark & - & \textbf{0.986}  & \textbf{0.498} \\  
    \midrule 
    INR $\rightarrow$ Large MLP & \Checkmark & - & 0.980 & 0.498 \\ 
    \bottomrule\bottomrule
    \end{tabular}
}
    \vspace{-0.8em}

   \label{tab:class_embedding}
\end{table}
\subsection{Ablation studies}

\noindent\textbf{Class-aware Embedding}. We conducted an investigation into the impact of class embedding methods and developed three class embedding methods based on the previously mentioned methods ~\cite{xu2022multi, dong2022category, xie2023category}. The first block of Table~\ref{tab:class_embedding} was used to compare the performance, while Figure~\ref{fig:class_embed} provided a schematic diagram of methods. We utilized UniAD without class information as the baseline. 
We discovered that directly concatenating class information with input features resulted in improved metrics, indicating the advantage of class information on network performance. While the three class-aware improvements vary significantly, they all effectively utilize the label as a prompt.

\noindent\textbf{Prompt Types.} We further compare the three prompt types mentioned in Figure \ref{fig:MINT}. 
Firstly, we can observe in the second block of Table~\ref{tab:class_embedding} that when using the label as a prompt, our performance is better than the previous three methods, indicating that the INR network we use is superior. Secondly, it can be noticed that using features from the classification network as prompts is the optimal choice because it provides finer-grained classification compared to the label. Using the image as a prompt does not use any label, but it still offers a certain performance improvement compared to UniAD due to the network advantage.  

\begin{table}[htb]
\centering
\scriptsize
\caption{Different submodule ablation study, \Checkmark represents the use of the corresponding operation.}
\label{tab:mint-ablation}
\small
\resizebox{0.6\linewidth}{!}{
\begin{tabular}{cccccc}
\toprule\toprule
$Adapter$ & $L_{CE}$ & $L_{Prior}$& $Query$ &I-AUROC & P-AUPR \\
\midrule
-& -& -& -& 0.971 & 0.447 \\
\Checkmark& - & -&  - &  0.975  & 0.445\\
\Checkmark& \Checkmark &  -& -  &  0.977  & 0.465\\
\Checkmark& \Checkmark  & - & \Checkmark  & 0.981  & 0.477 \\
\Checkmark& \Checkmark  & \Checkmark  & -& 0.983  & \textbf{0.502} \\

\Checkmark  &\Checkmark  & \Checkmark  & \Checkmark  & \textbf{0.986}  & 0.498  \\
\bottomrule\bottomrule
\end{tabular}
}
    \vspace{-0.8em}

\end{table} 

\noindent\textbf{Architecture Design}. The last block of Table~\ref{tab:class_embedding} shows the performance when replacing the INR network with a large MLP (from grid search). Despite having over 500 times more parameters than INR, MLP still struggles to achieve the mapping capability of INR.
Table~\ref{tab:mint-ablation} presented below displays the outcomes of ablation experiments conducted on each key operation of MINT-AD, including the adapter, the classification loss, the prior loss, and the class-aware embedding query. The results show that the feature adapter is useful for detection. It was observed that the classification loss had a negligible impact on performance. Further analysis of the impact of prior loss and class-aware embedding query revealed that both contributed to the improvement in performance. However, the inclusion of prior loss significantly enhanced pixel-level metrics by computing the mean and variance of the pixel distribution for the corresponding class, thereby aligning the reconstruction results with the distribution. Finally, combining all modules significantly improved image-level metrics, while pixel-level metrics slightly decreased compared to not using query embedding. This could be attributed to the class query focusing more on global class information, which may have a negative impact on pixel-level metrics. Refer to the appendix for further experimental outcomes, including computational complexity calculations and backbone selection. 

\section{Conclusion}
We investigate multi-class anomaly detection scenarios and propose a category-aware unified reconstruction network. It innovatively applies implicit neural representation in anomaly detection, mapping category encoding to feature space. To address the fitting of multiple samples within the same class, our model learns the category distribution by predicting the prior distribution and proposes a prior loss to update the network. Experimental results show that this paper's proposed MINT-AD algorithm outperforms existing algorithms on multiple datasets.

\noindent \textbf{Discussion.} In this work, we did not establish connections with relevant classes in anomaly detection. We believe that this is a reasonable direction for future improvements. At the same time, the security of the data and labels has not been fully studied. When applied in monitoring, it may be vulnerable to backdoor attacks, leading to privacy leaks. 



%
%

\bibliographystyle{splncs04}
\bibliography{CVPR/main}

\begin{thebibliography}{10}
\providecommand{\url}[1]{\texttt{#1}}
\providecommand{\urlprefix}{URL }
\providecommand{\doi}[1]{https://doi.org/#1}

\bibitem{bergmann2019mvtec}
Bergmann, P., Fauser, M., Sattlegger, D., Steger, C.: Mvtec ad--a comprehensive
  real-world dataset for unsupervised anomaly detection. In: Proceedings of the
  IEEE/CVF conference on computer vision and pattern recognition. pp.
  9592--9600 (2019)

\bibitem{bergmann2018improving}
Bergmann, P., L{\"o}we, S., Fauser, M., Sattlegger, D., Steger, C.: Improving
  unsupervised defect segmentation by applying structural similarity to
  autoencoders. arXiv preprint arXiv:1807.02011  (2018)

\bibitem{chen2021crossvit}
Chen, C.F.R., Fan, Q., Panda, R.: Crossvit: Cross-attention multi-scale vision
  transformer for image classification. In: Proceedings of the IEEE/CVF
  international conference on computer vision. pp. 357--366 (2021)

\bibitem{chen2022utrad}
Chen, L., You, Z., Zhang, N., Xi, J., Le, X.: Utrad: Anomaly detection and
  localization with u-transformer. Neural Networks  \textbf{147},  53--62
  (2022)

\bibitem{chen2021direct}
Chen, S., Wang, Z., Prisacariu, V.: Direct-posenet: Absolute pose regression
  with photometric consistency. In: 2021 International Conference on 3D Vision
  (3DV). pp. 1175--1185. IEEE (2021)

\bibitem{104}
Chen, Y., Tian, Y., Pang, G., Carneiro, G.: Deep one-class classification via
  interpolated gaussian descriptor (2021)

\bibitem{chen2022deep}
Chen, Y., Tian, Y., Pang, G., Carneiro, G.: Deep one-class classification via
  interpolated gaussian descriptor. In: Proceedings of the AAAI Conference on
  Artificial Intelligence. vol.~36, pp. 383--392 (2022)

\bibitem{cohen2020sub}
Cohen, N., Hoshen, Y.: Sub-image anomaly detection with deep pyramid
  correspondences. arXiv preprint arXiv:2005.02357  (2020)

\bibitem{defard2021padim}
Defard, T., Setkov, A., Loesch, A., Audigier, R.: Padim: a patch distribution
  modeling framework for anomaly detection and localization. In: International
  Conference on Pattern Recognition. pp. 475--489. Springer (2021)

\bibitem{ding2022catching}
Ding, C., Pang, G., Shen, C.: Catching both gray and black swans: Open-set
  supervised anomaly detection. In: Proceedings of the IEEE/CVF Conference on
  Computer Vision and Pattern Recognition. pp. 7388--7398 (2022)

\bibitem{dong2022category}
Dong, L., Li, Z., Xu, K., Zhang, Z., Yan, L., Zhong, S., Zou, X.:
  Category-aware transformer network for better human-object interaction
  detection. In: Proceedings of the IEEE/CVF Conference on Computer Vision and
  Pattern Recognition. pp. 19538--19547 (2022)

\bibitem{gudovskiy2022cflow}
Gudovskiy, D., Ishizaka, S., Kozuka, K.: Cflow-ad: Real-time unsupervised
  anomaly detection with localization via conditional normalizing flows. In:
  Proceedings of the IEEE/CVF Winter Conference on Applications of Computer
  Vision. pp. 98--107 (2022)

\bibitem{hao2022implicit}
Hao, Z., Mallya, A., Belongie, S., Liu, M.Y.: Implicit neural representations
  with levels-of-experts. Advances in Neural Information Processing Systems
  \textbf{35},  2564--2576 (2022)

\bibitem{jiang2022softpatch}
Jiang, X., Liu, J., Wang, J., Nie, Q., Wu, K., Liu, Y., Wang, C., Zheng, F.:
  Softpatch: Unsupervised anomaly detection with noisy data. Advances in Neural
  Information Processing Systems  \textbf{35},  15433--15445 (2022)

\bibitem{jiang2022survey}
Jiang, X., Xie, G., Wang, J., Liu, Y., Wang, C., Zheng, F., Jin, Y.: A survey
  of visual sensory anomaly detection. arXiv preprint arXiv:2202.07006  (2022)

\bibitem{jiang2023negative}
Jiang, X., Liu, F., Fang, Z., Chen, H., Liu, T., Zheng, F., Han, B.: Negative
  label guided ood detection with pretrained vision-language models. In: The
  Twelfth International Conference on Learning Representations (2023)

\bibitem{lee2022cfa}
Lee, S., Lee, S., Song, B.C.: Cfa: Coupled-hypersphere-based feature adaptation
  for target-oriented anomaly localization. IEEE Access  \textbf{10},
  78446--78454 (2022)

\bibitem{lee2023selformaly}
Lee, Y., Lim, H., Yoon, H.: Selformaly: Towards task-agnostic unified anomaly
  detection. arXiv preprint arXiv:2307.12540  (2023)

\bibitem{Lei_2023_CVPR}
Lei, J., Hu, X., Wang, Y., Liu, D.: Pyramidflow: High-resolution defect
  contrastive localization using pyramid normalizing flow. In: Proceedings of
  the IEEE/CVF Conference on Computer Vision and Pattern Recognition (CVPR).
  pp. 14143--14152 (June 2023)

\bibitem{li2021cutpaste}
Li, C.L., Sohn, K., Yoon, J., Pfister, T.: Cutpaste: Self-supervised learning
  for anomaly detection and localization. In: Proceedings of the IEEE/CVF
  Conference on Computer Vision and Pattern Recognition. pp. 9664--9674 (2021)

\bibitem{li2023efficient}
Li, H., Wu, J., Chen, H., Wang, M., Shen, C.: Efficient anomaly detection with
  budget annotation using semi-supervised residual transformer. arXiv preprint
  arXiv:2306.03492  (2023)

\bibitem{li2022towards}
Li, W., Zhan, J., Wang, J., Xia, B., Gao, B.B., Liu, J., Wang, C., Zheng, F.:
  Towards continual adaptation in industrial anomaly detection. In: Proceedings
  of the 30th ACM International Conference on Multimedia. pp. 2871--2880 (2022)

\bibitem{liu2024deep}
Liu, J., Xie, G., Wang, J., Li, S., Wang, C., Zheng, F., Jin, Y.: Deep
  industrial image anomaly detection: A survey. Machine Intelligence Research
  \textbf{21}(1),  104--135 (2024)

\bibitem{liu2023deep}
Liu, J., Xie, G., Wang, J., Li, S., Wang, C., Zheng, F., Jin, Y.: Deep visual
  anomaly detection in industrial manufacturing: A survey. arXiv preprint
  arXiv:2301.11514  (2023)

\bibitem{liu2020towards}
Liu, W., Li, R., Zheng, M., Karanam, S., Wu, Z., Bhanu, B., Radke, R.J., Camps,
  O.: Towards visually explaining variational autoencoders. In: Proceedings of
  the IEEE/CVF Conference on Computer Vision and Pattern Recognition. pp.
  8642--8651 (2020)

\bibitem{liu2023diversity}
Liu, W., Chang, H., Ma, B., Shan, S., Chen, X.: Diversity-measurable anomaly
  detection. arXiv preprint arXiv:2303.05047  (2023)

\bibitem{liu2023simplenet}
Liu, Z., Zhou, Y., Xu, Y., Wang, Z.: Simplenet: A simple network for image
  anomaly detection and localization. arXiv preprint arXiv:2303.15140  (2023)

\bibitem{Lu2023RemovingAA}
Lu, F., Yao, X., Fu, C.W., Jia, J.: Removing anomalies as noises for industrial
  defect localization. 2023 IEEE/CVF International Conference on Computer
  Vision (ICCV) pp. 16120--16129 (2023)

\bibitem{Lu2023HierarchicalVQ}
Lu, R., Wu, Y., Tian, L., Wang, D., Chen, B., Liu, X., Hu, R.: Hierarchical
  vector quantized transformer for multi-class unsupervised anomaly detection.
  NeurIPS  (2023)

\bibitem{mehta2021modulated}
Mehta, I., Gharbi, M., Barnes, C., Shechtman, E., Ramamoorthi, R., Chandraker,
  M.: Modulated periodic activations for generalizable local functional
  representations. In: Proceedings of the IEEE/CVF International Conference on
  Computer Vision. pp. 14214--14223 (2021)

\bibitem{mildenhall2021nerf}
Mildenhall, B., Srinivasan, P.P., Tancik, M., Barron, J.T., Ramamoorthi, R.,
  Ng, R.: Nerf: Representing scenes as neural radiance fields for view
  synthesis. Communications of the ACM  \textbf{65}(1),  99--106 (2021)

\bibitem{mishra2021vt}
Mishra, P., Verk, R., Fornasier, D., Piciarelli, C., Foresti, G.L.: Vt-adl: A
  vision transformer network for image anomaly detection and localization. In:
  2021 IEEE 30th International Symposium on Industrial Electronics (ISIE). pp.
  01--06. IEEE (2021)

\bibitem{neal1998view}
Neal, R.M., Hinton, G.E.: A view of the em algorithm that justifies
  incremental, sparse, and other variants. Learning in graphical models pp.
  355--368 (1998)

\bibitem{park2019deepsdf}
Park, J.J., Florence, P., Straub, J., Newcombe, R., Lovegrove, S.: Deepsdf:
  Learning continuous signed distance functions for shape representation. In:
  Proceedings of the IEEE/CVF conference on computer vision and pattern
  recognition. pp. 165--174 (2019)

\bibitem{qiu2022latent}
Qiu, C., Li, A., Kloft, M., Rudolph, M., Mandt, S.: Latent outlier exposure for
  anomaly detection with contaminated data. In: International Conference on
  Machine Learning. pp. 18153--18167. PMLR (2022)

\bibitem{radford2021learning}
Radford, A., Kim, J.W., Hallacy, C., Ramesh, A., Goh, G., Agarwal, S., Sastry,
  G., Askell, A., Mishkin, P., Clark, J., et~al.: Learning transferable visual
  models from natural language supervision. In: International conference on
  machine learning. pp. 8748--8763. PMLR (2021)

\bibitem{Reiss2021PANDAAP}
Reiss, T., Cohen, N., Bergman, L., Hoshen, Y.: Panda: Adapting pretrained
  features for anomaly detection and segmentation. 2021 IEEE/CVF Conference on
  Computer Vision and Pattern Recognition (CVPR) pp. 2805--2813 (2021)

\bibitem{rombach2022high}
Rombach, R., Blattmann, A., Lorenz, D., Esser, P., Ommer, B.: High-resolution
  image synthesis with latent diffusion models. In: Proceedings of the IEEE/CVF
  conference on computer vision and pattern recognition. pp. 10684--10695
  (2022)

\bibitem{roth2021towards}
Roth, K., Pemula, L., Zepeda, J., Sch{\"o}lkopf, B., Brox, T., Gehler, P.:
  Towards total recall in industrial anomaly detection. arXiv preprint
  arXiv:2106.08265  (2021)

\bibitem{sabokrou2018adversarially}
Sabokrou, M., Khalooei, M., Fathy, M., Adeli, E.: Adversarially learned
  one-class classifier for novelty detection. In: Proceedings of the IEEE
  conference on computer vision and pattern recognition. pp. 3379--3388 (2018)

\bibitem{salehi2021multiresolution}
Salehi, M., Sadjadi, N., Baselizadeh, S., Rohban, M.H., Rabiee, H.R.:
  Multiresolution knowledge distillation for anomaly detection. In: Proceedings
  of the IEEE/CVF conference on computer vision and pattern recognition. pp.
  14902--14912 (2021)

\bibitem{VincentSitzmann2020ImplicitNR}
Sitzmann, V., Martel, J.N.P., Bergman, A.W., Lindell, D.B., Wetzstein, G.:
  Implicit neural representations with periodic activation functions. Neural
  Information Processing Systems  (2020)

\bibitem{tan2019efficientnet}
Tan, M., Le, Q.: Efficientnet: Rethinking model scaling for convolutional
  neural networks. In: International conference on machine learning. pp.
  6105--6114. PMLR (2019)

\bibitem{xie2023category}
Xie, C., Zeng, F., Hu, Y., Liang, S., Wei, Y.: Category query learning for
  human-object interaction classification. arXiv preprint arXiv:2303.14005
  (2023)

\bibitem{xie2023iad}
Xie, G., Wang, J., Liu, J., Lyu, J., Liu, Y., Wang, C., Zheng, F., Jin, Y.:
  Im-iad: Industrial image anomaly detection benchmark in manufacturing. arXiv
  preprint arXiv:2301.13359  (2023)

\bibitem{xu2022multi}
Xu, L., Ouyang, W., Bennamoun, M., Boussaid, F., Xu, D.: Multi-class token
  transformer for weakly supervised semantic segmentation. In: Proceedings of
  the IEEE/CVF Conference on Computer Vision and Pattern Recognition. pp.
  4310--4319 (2022)

\bibitem{yang2022openood}
Yang, J., Wang, P., et~al.: Openood: Benchmarking generalized
  out-of-distribution detection. Advances in Neural Information Processing
  Systems  \textbf{35},  32598--32611 (2022)

\bibitem{yang2021generalized}
Yang, J., Zhou, K., Li, Y., Liu, Z.: Generalized out-of-distribution detection:
  A survey. arXiv preprint arXiv:2110.11334  (2021)

\bibitem{yao2022explicit}
Yao, X., Zhang, C., Li, R.: Explicit boundary guided semi-push-pull contrastive
  learning for better anomaly detection. arXiv preprint arXiv:2207.01463
  (2022)

\bibitem{yi2020patch}
Yi, J., Yoon, S.: Patch svdd: Patch-level svdd for anomaly detection and
  segmentation. In: Proceedings of the Asian Conference on Computer Vision
  (2020)

\bibitem{you2022unified}
You, Z., Cui, L., Shen, Y., Yang, K., Lu, X., Zheng, Y., Le, X.: A unified
  model for multi-class anomaly detection. arXiv preprint arXiv:2206.03687
  (2022)

\bibitem{yuan2023devil}
Yuan, M., Xia, Y., Dong, H., Chen, Z., Yao, J., Qiu, M., Yan, K., Yin, X., Shi,
  Y., Chen, X., et~al.: Devil is in the queries: Advancing mask transformers
  for real-world medical image segmentation and out-of-distribution
  localization. arXiv preprint arXiv:2304.00212  (2023)

\bibitem{7}
Zavrtanik, V., Kristan, M., Sko{\v{c}}aj, D.: Draem-a discriminatively trained
  reconstruction embedding for surface anomaly detection. In: Proceedings of
  the IEEE/CVF International Conference on Computer Vision. pp. 8330--8339
  (2021)

\bibitem{Zhao_2023_CVPR}
Zhao, Y.: Omnial: A unified cnn framework for unsupervised anomaly
  localization. In: Proceedings of the IEEE/CVF Conference on Computer Vision
  and Pattern Recognition (CVPR). pp. 3924--3933 (June 2023)

\bibitem{zou2022spot}
Zou, Y., Jeong, J., Pemula, L., Zhang, D., Dabeer, O.: Spot-the-difference
  self-supervised pre-training for anomaly detection and segmentation. In:
  Computer Vision--ECCV 2022: 17th European Conference, Tel Aviv, Israel,
  October 23--27, 2022, Proceedings. pp. 392--408. Springer (2022)

\end{thebibliography}

\clearpage
\setcounter{page}{1}
\appendix
\section{Discussions about Unified Setting}

\subsection{Comparing with Ensemble Separated Models}

In this paper, we contend that category information is readily accessible within real-world industrial production processes. When it comes to multi-class anomaly detection tasks, the ability to distinguish categories can also be achieved through classifier training. Therefore, in some simple scenarios, the effect of a unified model can be achieved by integrating multiple independent models. For example, we can modify SOTA single-class approaches such as PatchCore for the unified setting by i) using a classifier to determine the image's category, and ii) utilizing a gallery of normal images from the relevant category alone. As PatchCore (and improved versions) perform better than this approach in the single-class setting, it is highly likely to outperform. However, it is worth noting that this approach may bring about several challenges. For instance, it requires a significant amount of memory to store models for different categories (as we have trained a unified model for even 30 categories), and in practical usage, loading different models for detection increases complexity. 

Furthermore, researching unified training tasks not only follows the previous ``one-for-all'' scheme~\cite{Lu2023HierarchicalVQ, you2022unified, Zhao_2023_CVPR} but also contributes to enhancing the overall capabilities of the model. On the one hand, the data within individual categories in existing classifications can further be subdivided into multiple subcategories. Therefore, it is significant to research how to address multi-class detection problems with a unified model. On the other hand, tackling multi-class detection tasks may also lead to improvements in one-class tasks. In Section~\ref{compare-seperate}, we delve into the reasons behind MINT-AD's superior performance in multi-class scenarios compared to single-class ones.

\begin{figure*}[hb] 
    \centering
    \includegraphics[width=\linewidth]{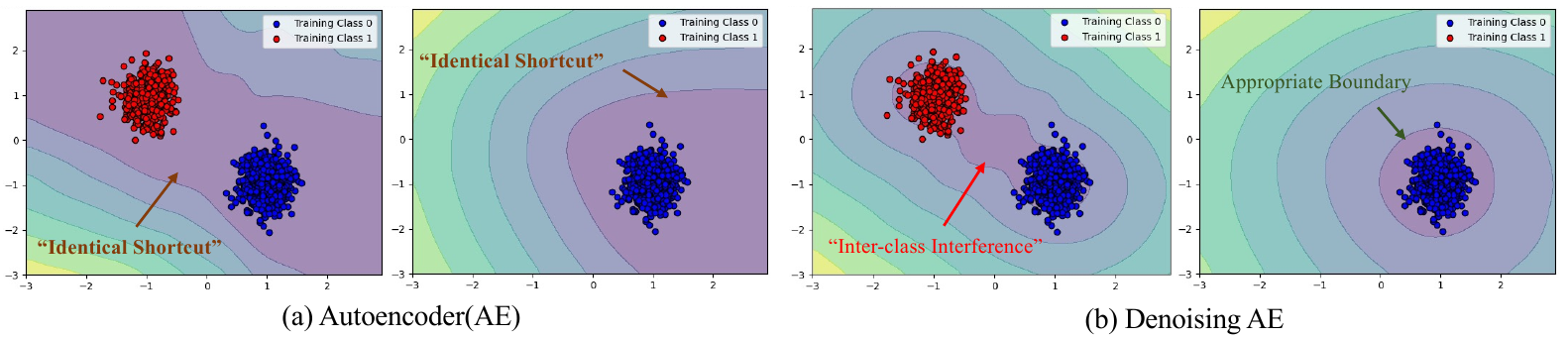}
    \caption{
    Decision boundaries on 2D synthetic multi-class data with different models. (a) vanilla MLP-based autoencoder, the ``identical shortcut'' occurs even during training with a single cluster. (b) denoising AE, where the reconstruction task is translated into a denoising task. The ``shortcut'' problem is solved, but ``inter-class interference'' occurs in a unified training setting.}
    \label{fig:shortcut}
\end{figure*}

\subsection{Using Category Information}
In the current setup of unified anomaly detection, the object categories are not within the scope of detection. The model needs to differentiate between normal and anomalous samples within each category rather than detecting semantic categories as in semantic anomaly detection~\cite{yang2021generalized, jiang2022survey}. In practical applications, category information is readily available. For example, in industrial quality inspection scenarios mentioned in the paper, data collection may involve multiple production lines or cameras installed at different viewpoints on the same production line, indicating that the data is inherently classifiable.

Existing unified anomaly detection methods~\cite{you2022unified, Zhao_2023_CVPR, Lu2023HierarchicalVQ} do not leverage category information. However, category information can assist the model, at least during training. Therefore, in our proposed approach MINT-AD, we incorporate the object's category labels only during the training phase. Additionally, we explore the effects of different approaches in utilizing category information. 

Overall, in our setting, the category information does not compromise the integrity of the detection process. In practical scenarios, where category information is easily accessible, it can serve as an additional advantage in facilitating the classification and organization of data.

\subsection{``Inter-class Interference''}
In previous research~\cite{you2022unified},  ``identical shortcut'' is regarded as the main problem introduced by the unified setting. However, the issue of ``identical shortcuts'' is actually more related to the model itself rather than the specific task at hand. As shown in the toy dataset example in Figure~\ref{fig:shortcut}, the vanilla autoencoder exhibits the ``shortcut'' problem in both single- and multi-class training scenarios.  The ``identical shortcut'' problem occurring in simple one-component, indicates that the current training method is not sufficient to allow the model to reasonably depict normal data.
On the other hand, the denoising autoencoder, which adds a jitter into input during training, does not suffer from this problem in the single-class scenario. However, there is mutual interference between the two classes' training data when it comes to multi-class training. Therefore, we believe this influence is not solely attributable to the model's ``shortcut'' but can be defined as ``inter-class interference''. 

\begin{figure}[htb]
    \centering
    \includegraphics[width=0.7\linewidth]{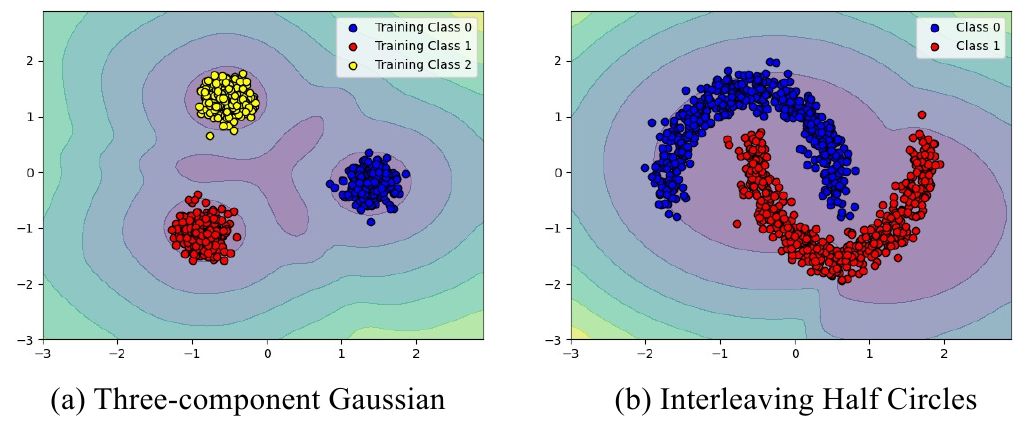}
    \caption{
    ``Inter-class interference'' observed on other datasets.}
    \label{fig:other-toy-dataset}
    \vspace{-0.8em}
\end{figure}

\begin{figure}[htb]
    \vspace{-0.8em}

    \centering
    \includegraphics[width=0.8\linewidth]{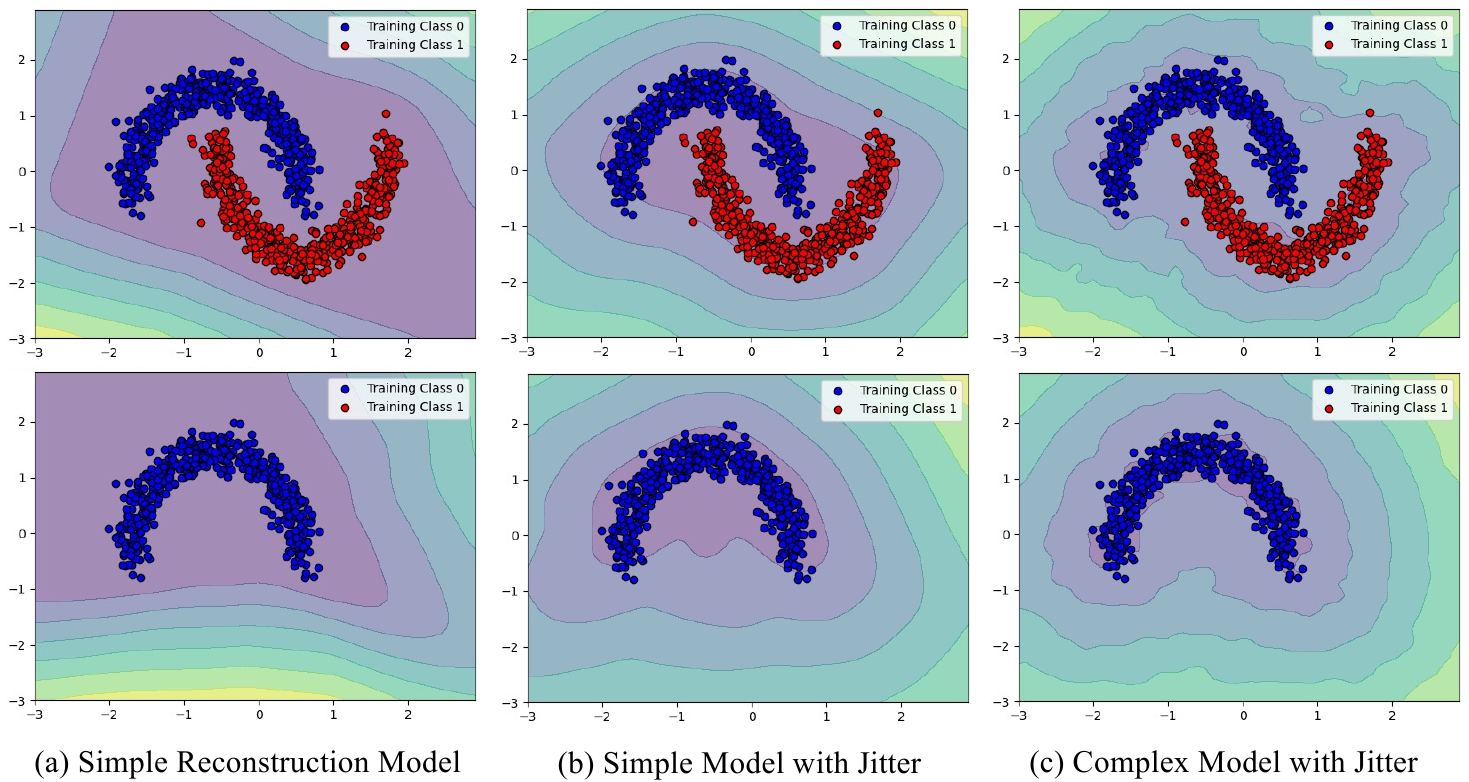}
    \caption{
    The relationship between decision boundaries and models' complexity.}
    \label{fig:toy-complex-model}
    \vspace{-0.8em}
    
\end{figure}

In addition, we conducted more experiments and visualizations to demonstrate that ``inter-class interference'' is an inevitable phenomenon when modeling multi-class data with a single model. Figure~\ref{fig:other-toy-dataset} shows the decision boundaries of denoising AE in two other multi-class shapes, demonstrating the universality of ``inter-class interference''. 
In Figure~\ref{fig:toy-complex-model}, we tried models with different complexities and found that using a model with higher complexity can solve the problem of ``inter-class interference'' to some extent, but it also brings the problem of non-smooth decision boundaries. 
Therefore, we can conclude that the method proposed by UniAD to add jitter in the reconstruction network is indeed a more effective training method for solving ``identical shortcut'', but it does not handle the ``inter-class interference'' unique to multi-class tasks. 

\section{More Results}
\subsection{Computational Complexity}
We calculate the inference FLOPs and parameters of MINT-AD and other baselines in Table \ref{tab-complexity}. The computational complexity is calculated with input images uniformly resized to 224 × 224. The results show that while MINT-AD has slightly higher computational requirements than UniAD, the complexity remains relatively low. Cause we use EffienctNet as the backbone, the FLOPs and Parameters are small. Compared to other one-class methods, the computational advantages of unified models become more pronounced, especially when increasing the batch size. In this test, we used Batchsize as the size of the training set, and it can be observed that methods like PatchCore, which relies on memory, exhibit exponential increases in inference time as the dataset size grows.


\begin{table}[h]
\caption{Complexity comparison between our MINT-AD and other baselines. }
\label{tab-complexity}
\centering
\setlength\tabcolsep{3pt}
\resizebox{\linewidth}{!}{
\begin{tabular}{l|c|cccccc}
\toprule\toprule
Methods      & Batchsize & PaDiM~\cite{defard2021padim} & PatchCore~\cite{roth2021towards} & DREAM~\cite{7}  & SimpleNet~\cite{liu2023simplenet} & UniAD~\cite{you2022unified}  & Ours \\ \midrule 
FLOPs (G)  & 1 & 11.46 & 36.96     & 152.10 & 47.06 & 0.92      & 1.13 \\
Params (M) & 1 & 68.88  & 49.73     & 97.42  & 139.34 & 6.27     & 9.44\\ \midrule 
Inference Time (s) & 1 & 21.368 & 0.035 & 0.02  & 0.014 & 0.032 & 0.026 \\ 
&8 & 28.729 & 1.191 & 0.14 & 0.06 & 0.068  & 0.035 \\ 
&16 & 30.475 & 4.734 & 0.218 & 0.112 & 0.089  & 0.057 \\ 
&32 & 31.197 & 18.33 & 0.435 & 0.221 & 0.146  & 0.104 \\ 
&64 & 33.121 & 72.204 & 0.873 & 0.435 & 0.2305  & 0.194 \\ 
\bottomrule\bottomrule
\end{tabular}}
    \vspace{-0.8em}

\end{table}

\subsection{Qualitative Results}
\begin{figure*}[htb]
    \centering
    \resizebox{\linewidth}{!}{
    \includegraphics[width=0.058\textwidth]{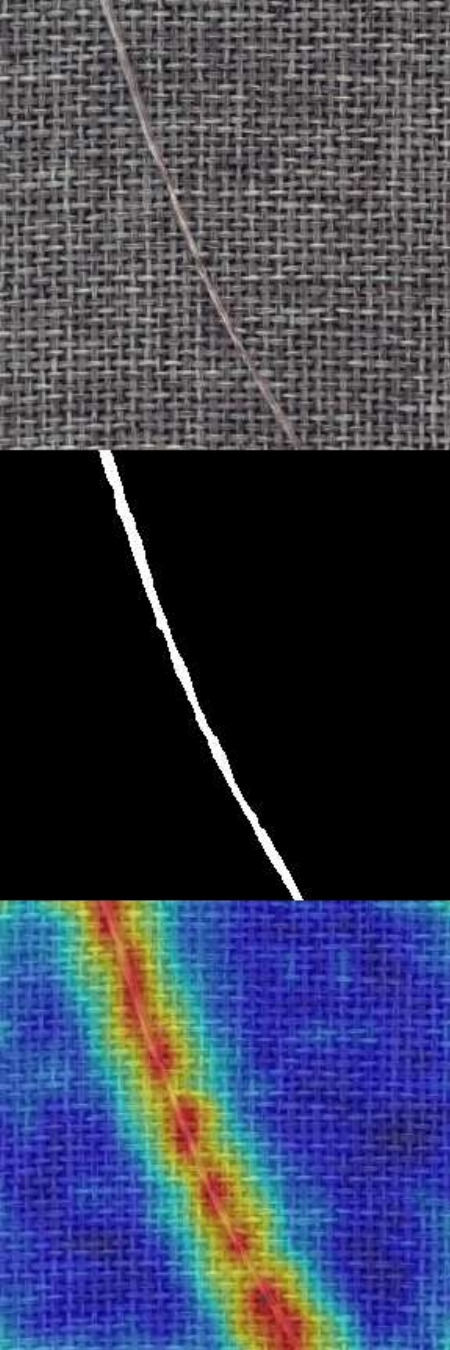}
    \includegraphics[width=0.058\textwidth]{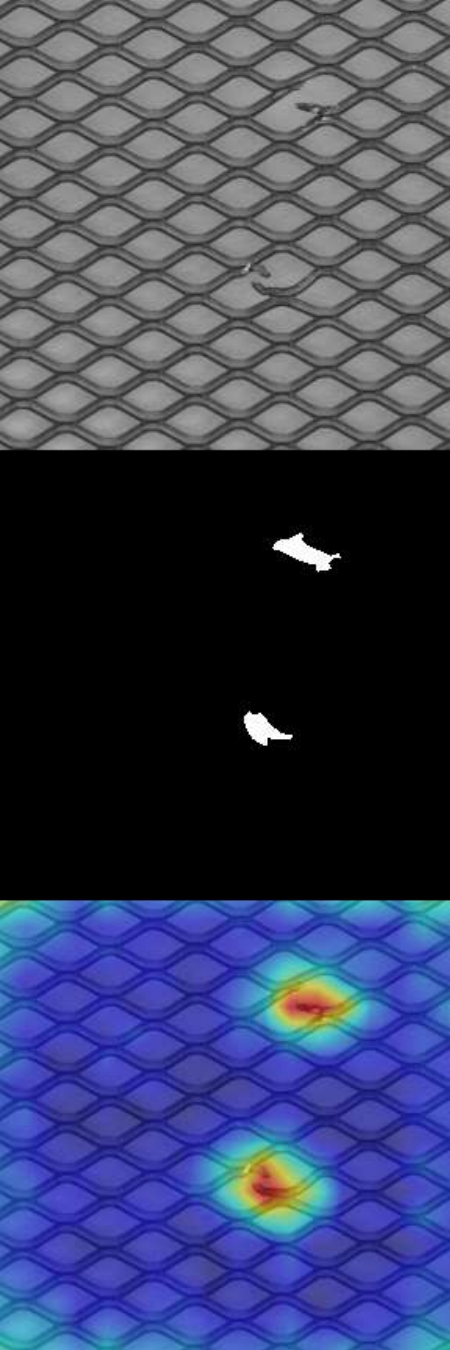}
    \includegraphics[width=0.058\textwidth]{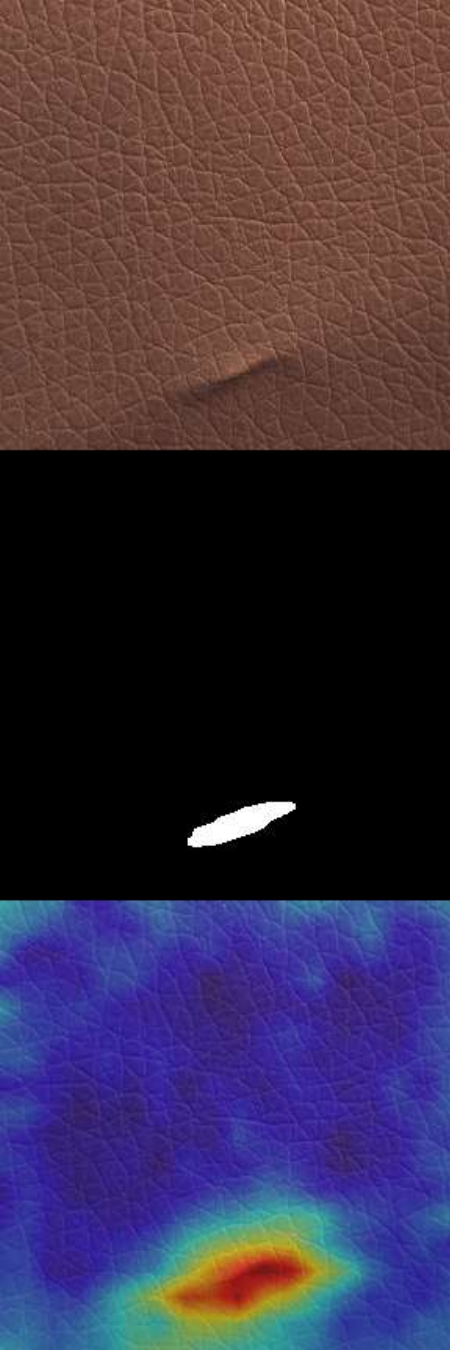}
    \includegraphics[width=0.058\textwidth]{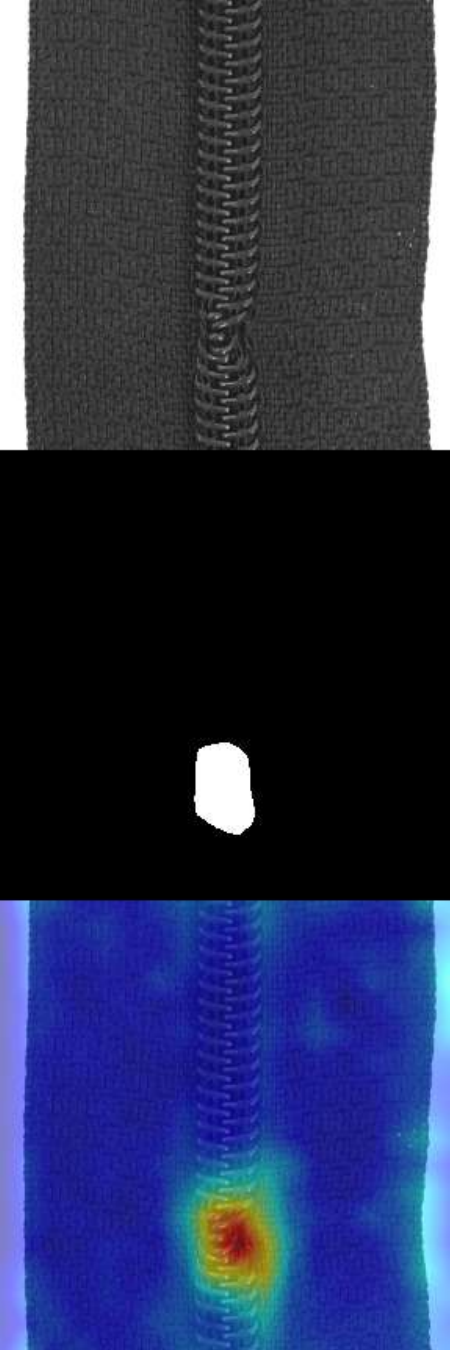}
    \includegraphics[width=0.058\textwidth]{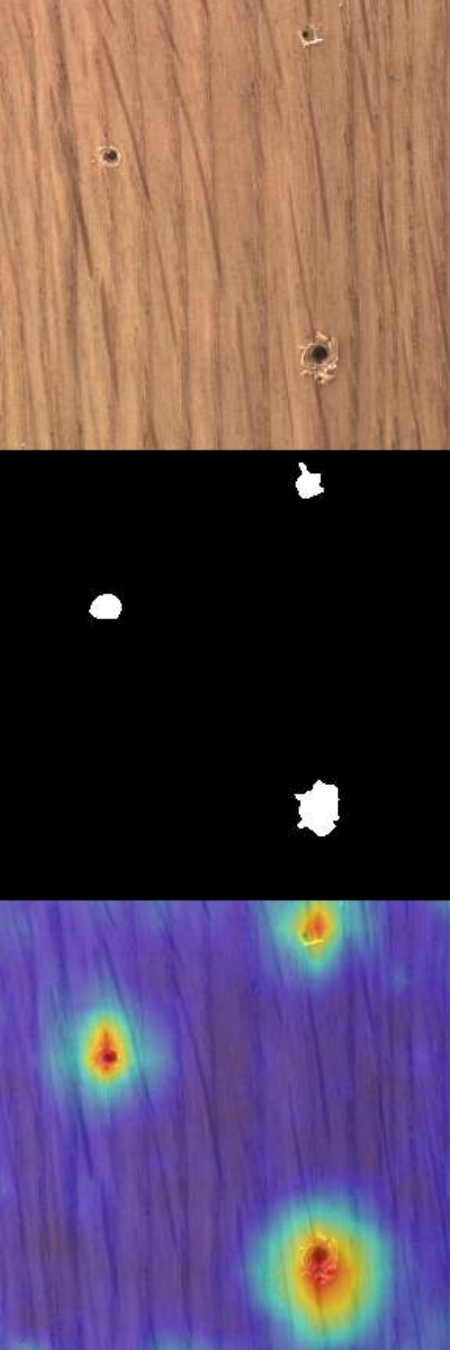}
    \includegraphics[width=0.058\textwidth]{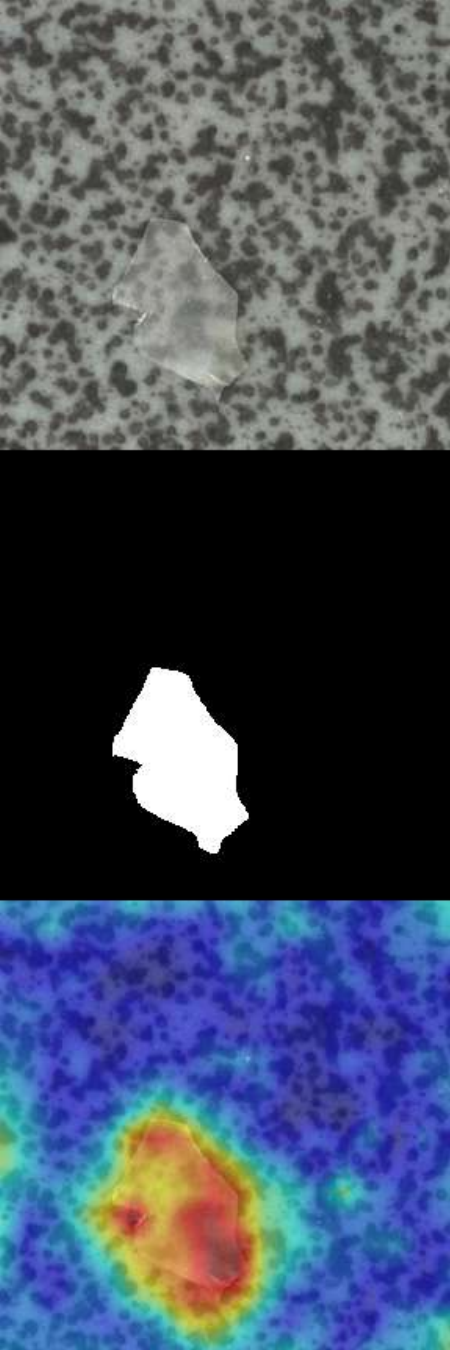}
    \includegraphics[width=0.058\textwidth]{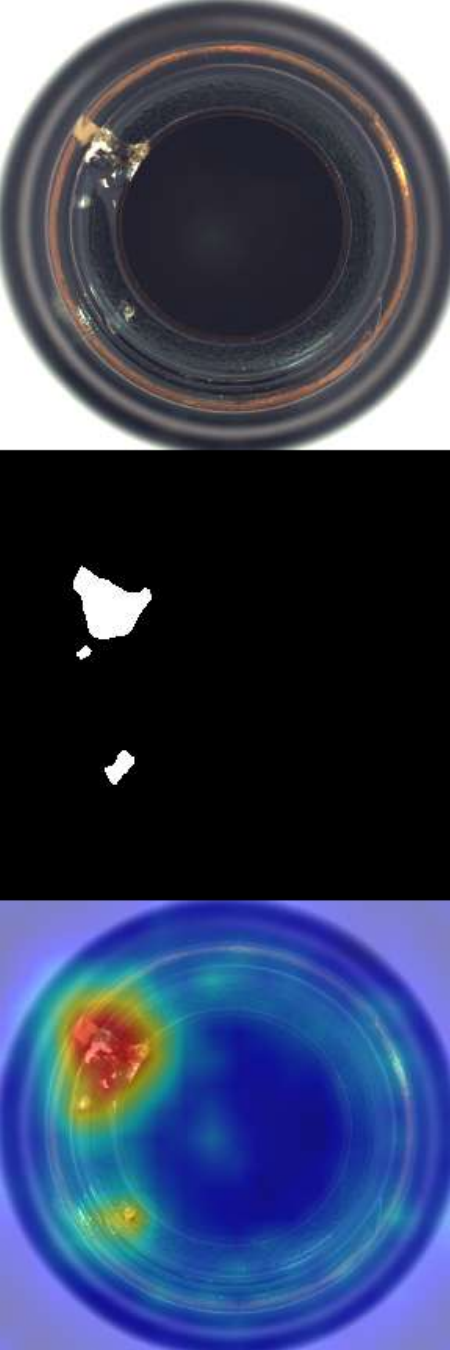} 
    \includegraphics[width=0.058\textwidth]{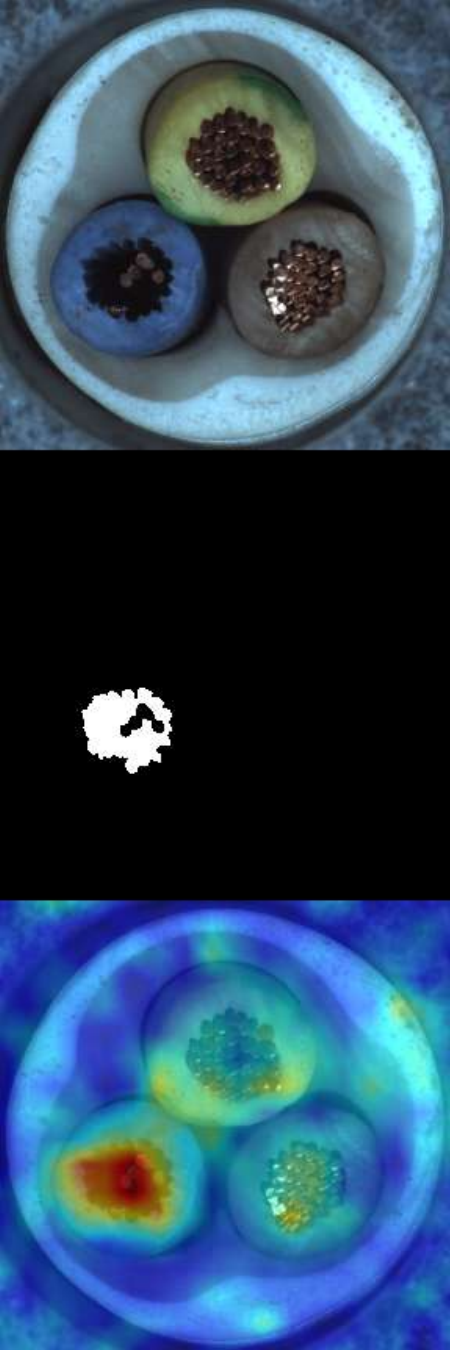}
    }\\
    \resizebox{0.9\linewidth}{!}{
    \includegraphics[width=0.058\textwidth]{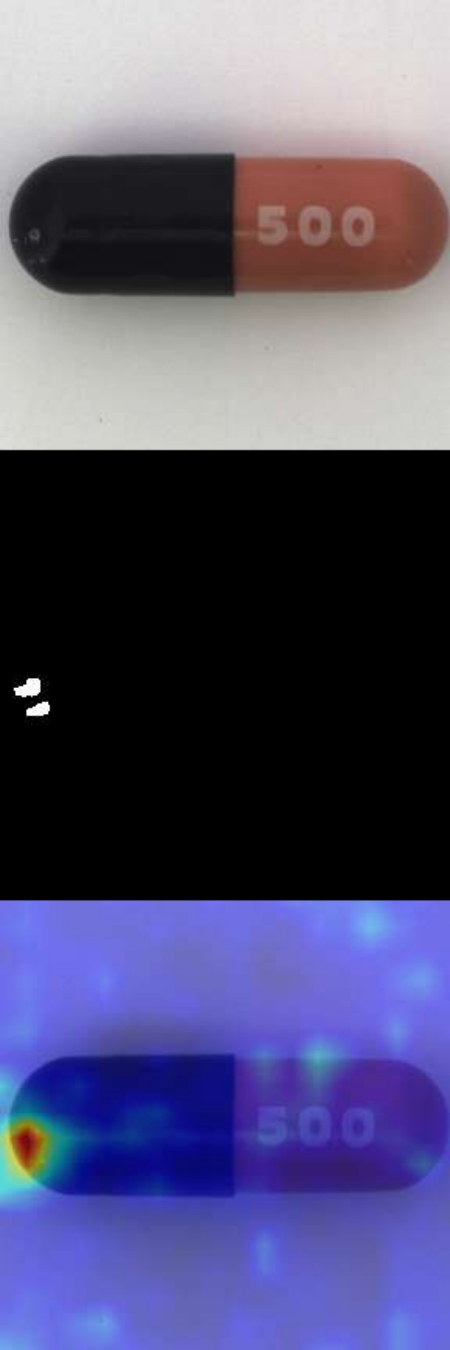}
    \includegraphics[width=0.058\textwidth]{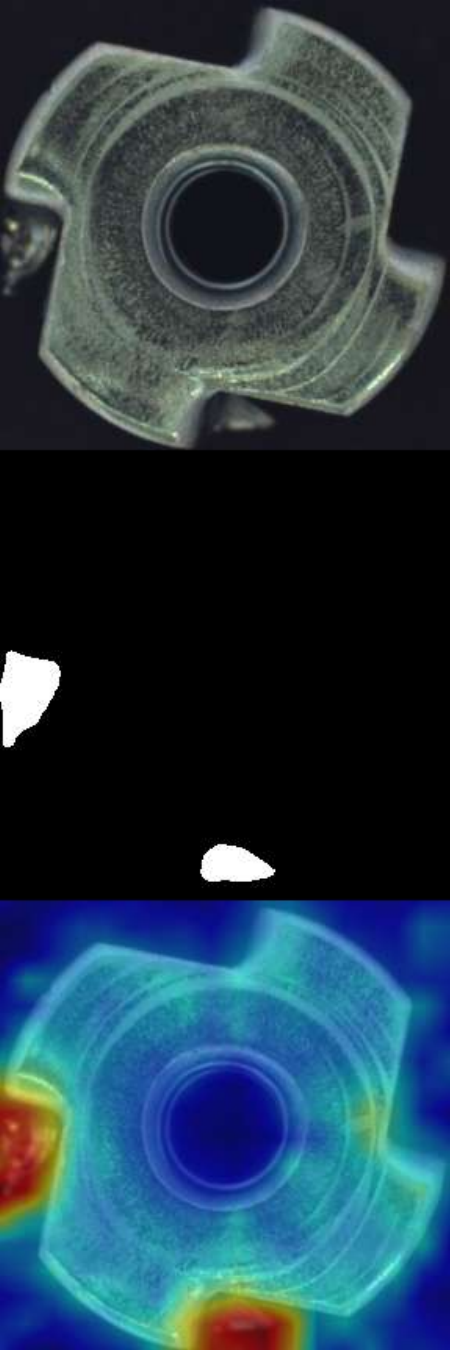}
    \includegraphics[width=0.058\textwidth]{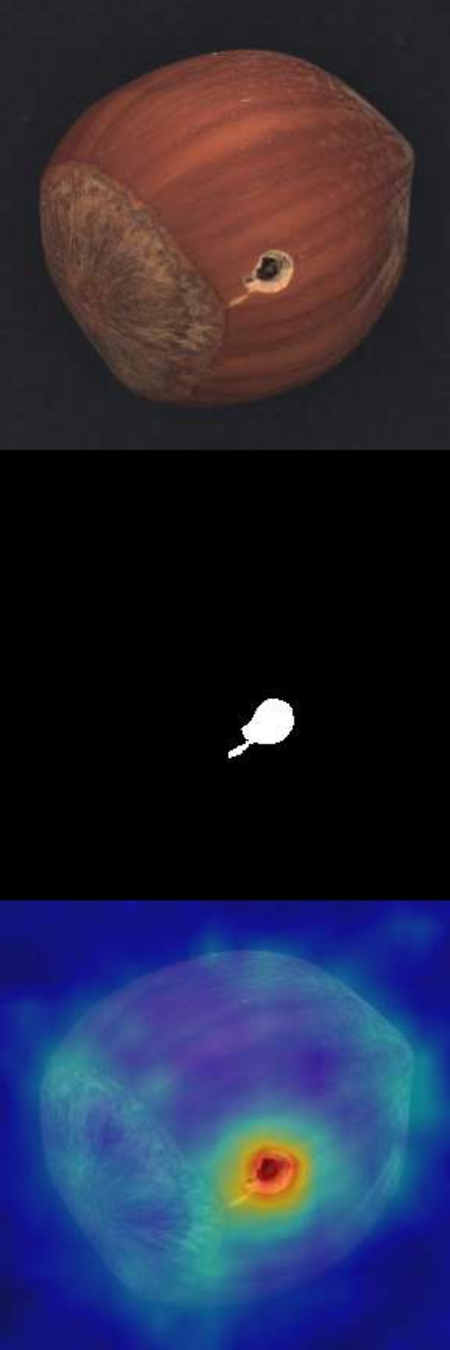}
    \includegraphics[width=0.058\textwidth]{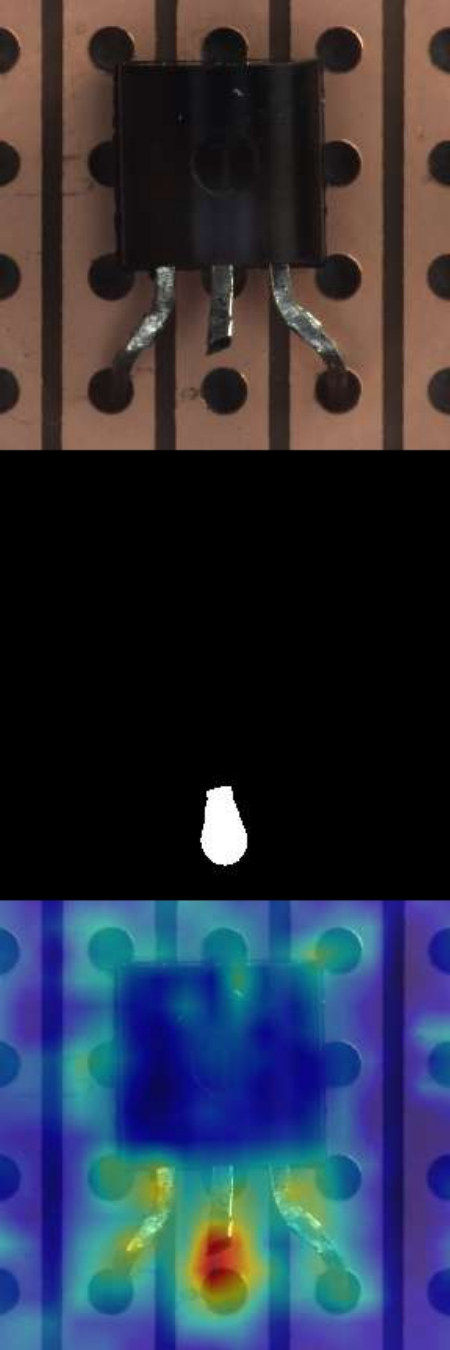}
    \includegraphics[width=0.058\textwidth]{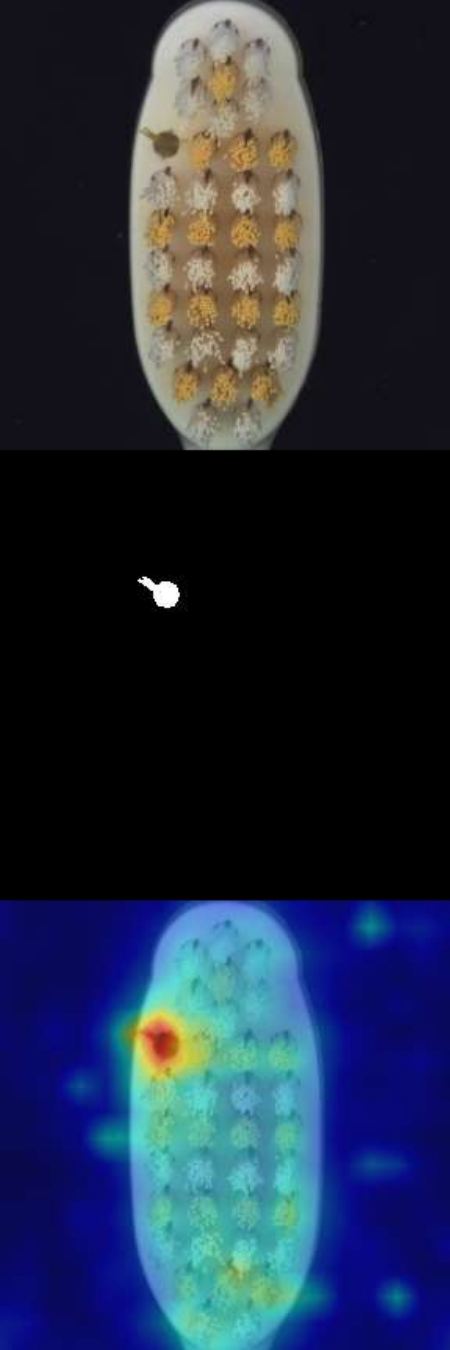}
    \includegraphics[width=0.058\textwidth]{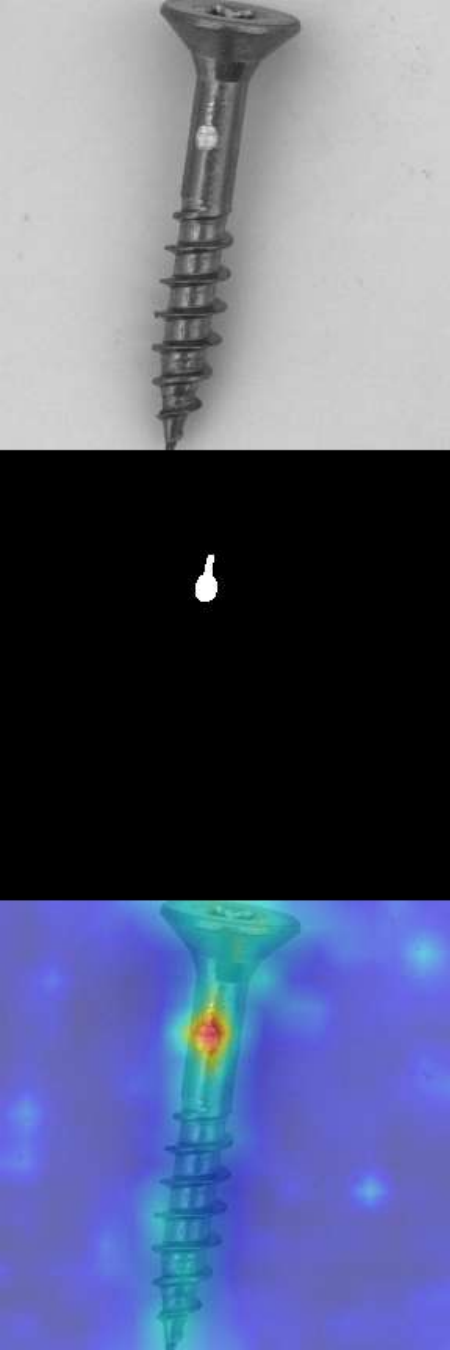}
    \includegraphics[width=0.058\textwidth]{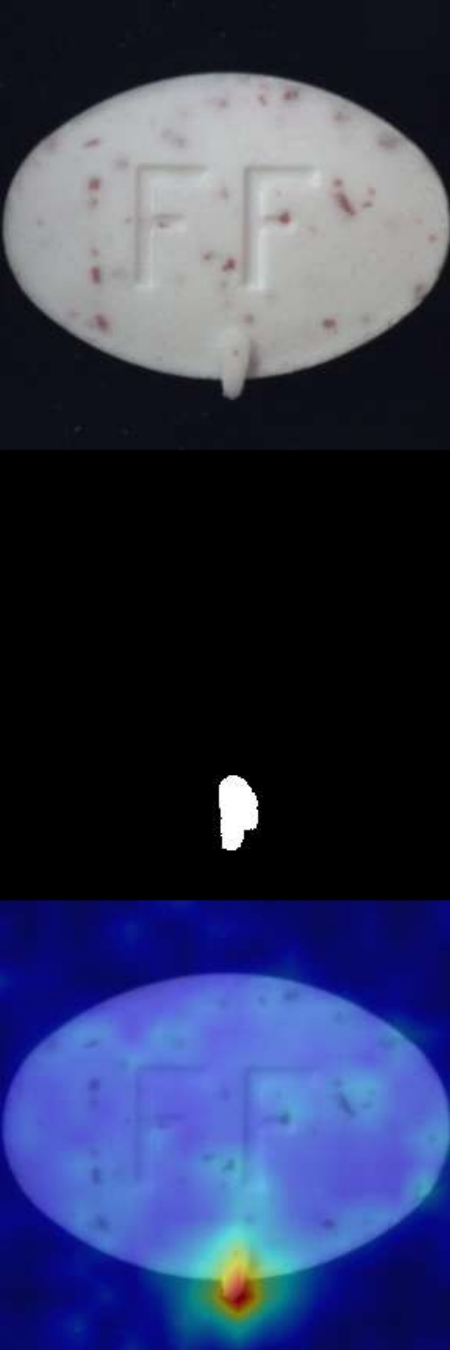}
    }
    \caption{Qualitative results, where sampled image, ground truth, and anomaly map are shown in sequence for each class in MVTec AD.}
    \label{mvtec_figure}
    \vspace{-0.8em}
    
\end{figure*}

\begin{figure*}[htb]
    \centering
    \resizebox{0.8\linewidth}{!}{
    \includegraphics[width=0.075\textwidth]{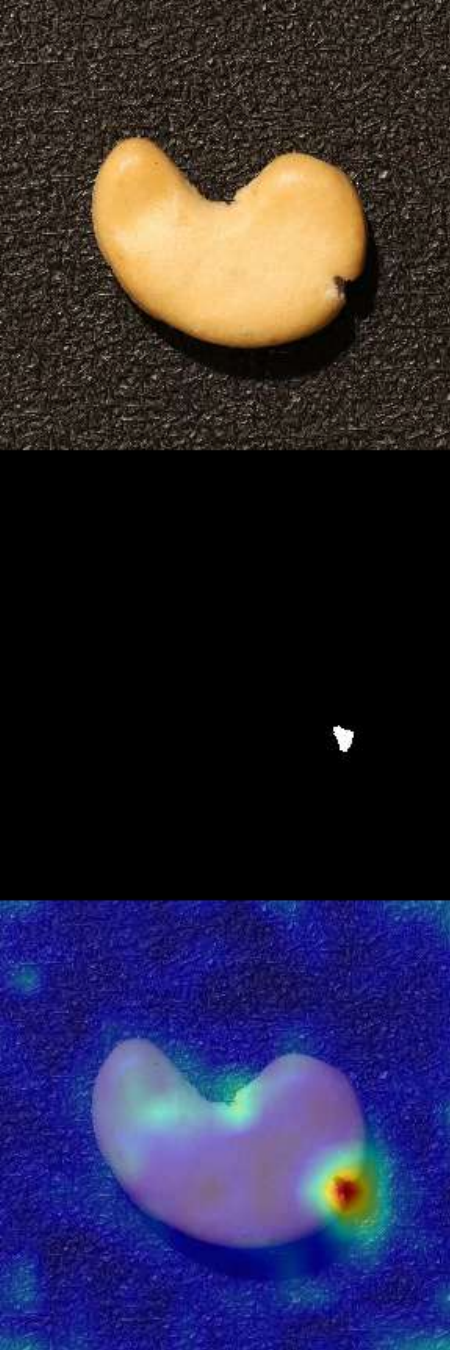}
    \includegraphics[width=0.075\textwidth]{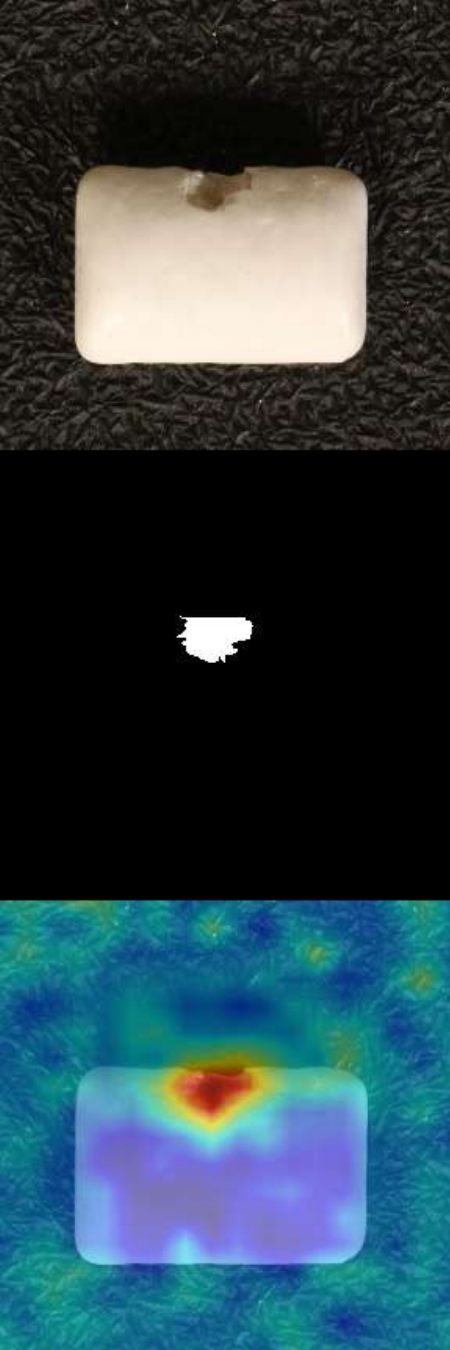}
    \includegraphics[width=0.075\textwidth]{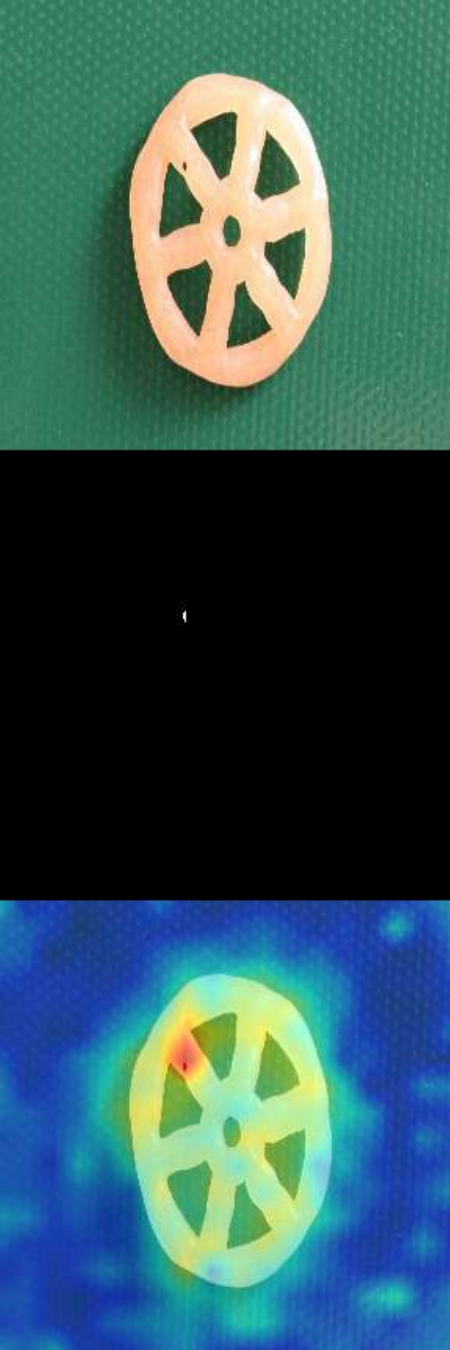}
    \includegraphics[width=0.075\textwidth]{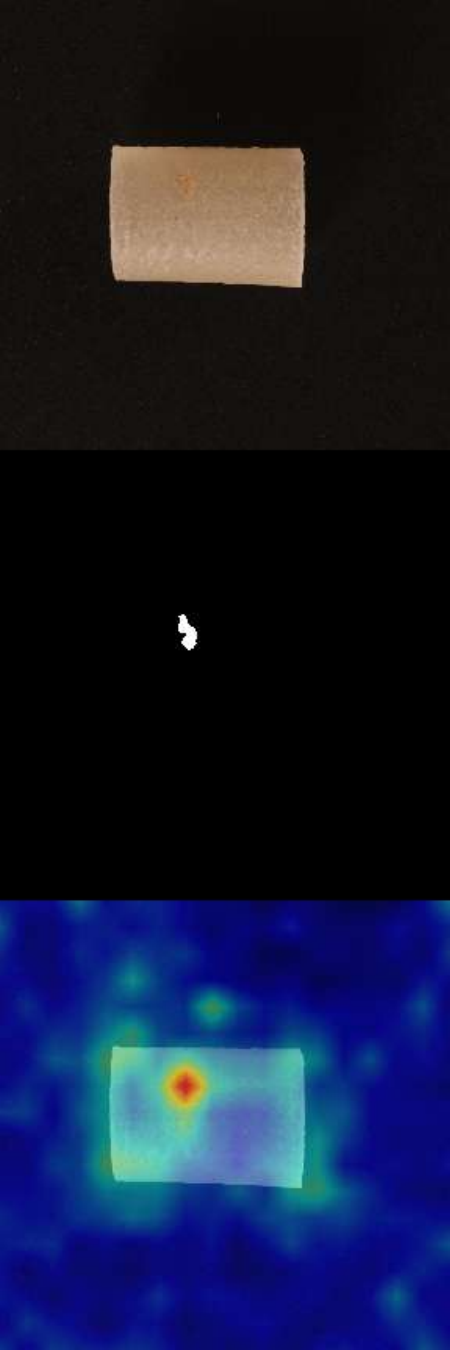}
    \includegraphics[width=0.075\textwidth]{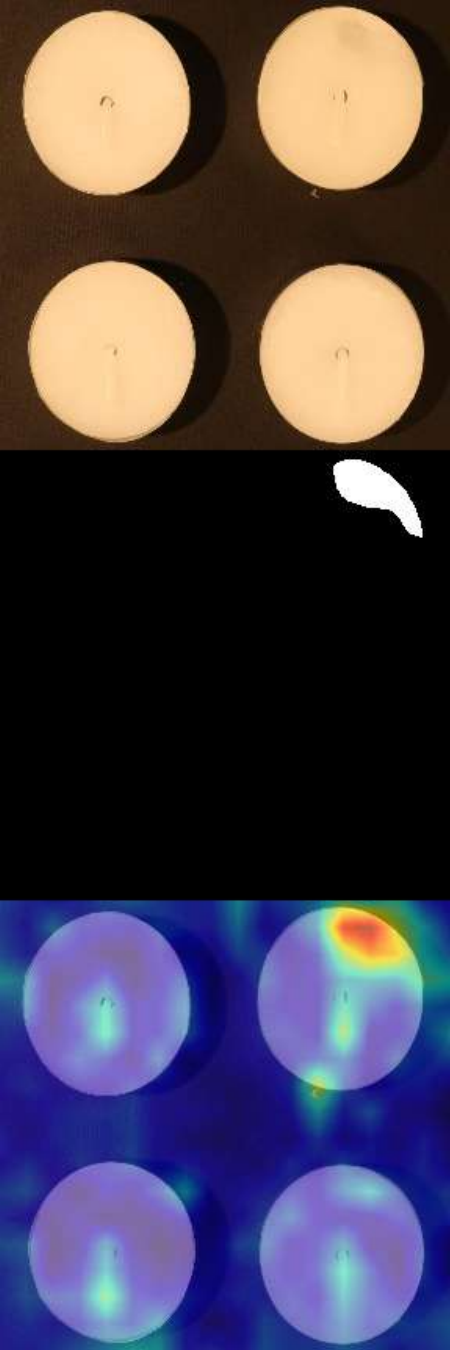}
    \includegraphics[width=0.075\textwidth]{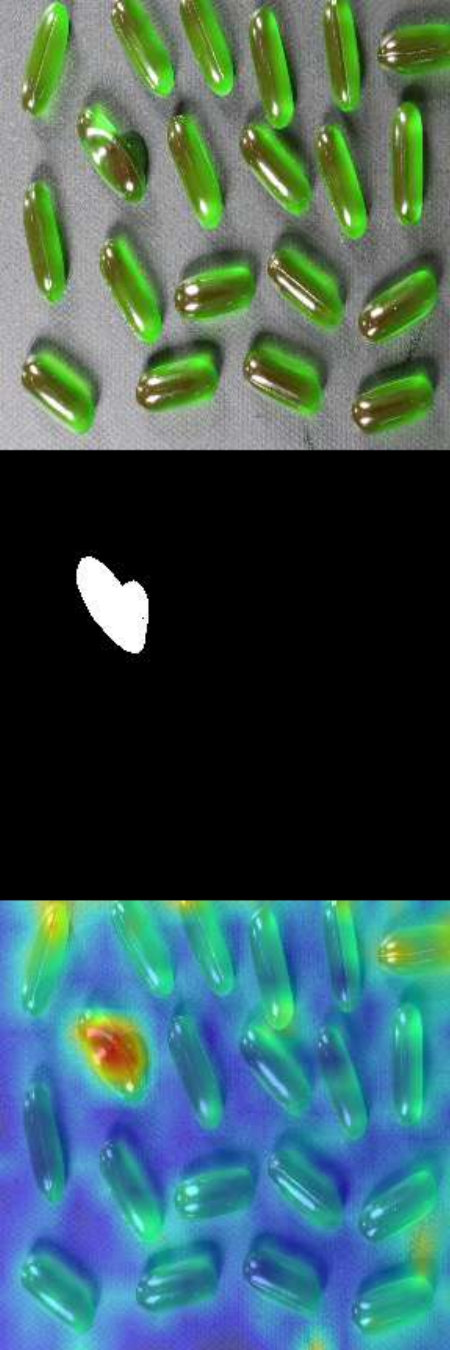}
    }\\
    \resizebox{0.8\linewidth}{!}{
    \includegraphics[width=0.075\textwidth]{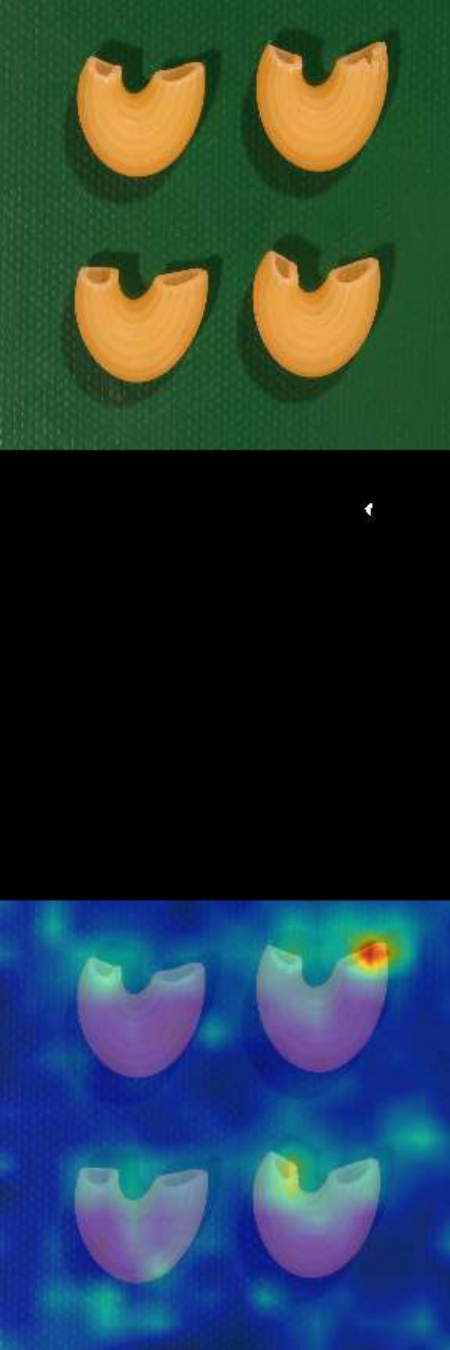}
    \includegraphics[width=0.075\textwidth]{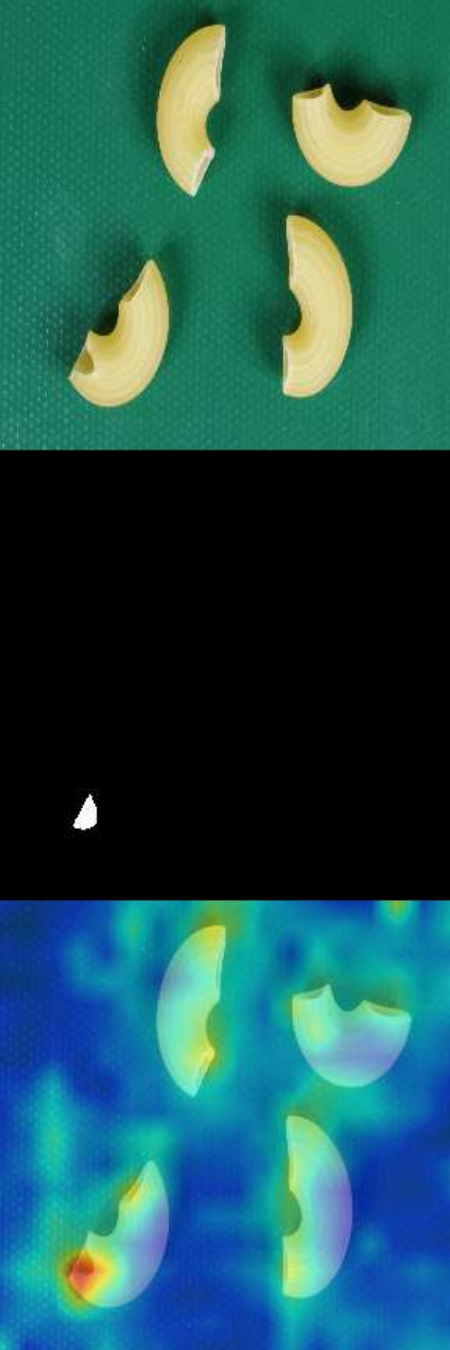}
    \includegraphics[width=0.075\textwidth]{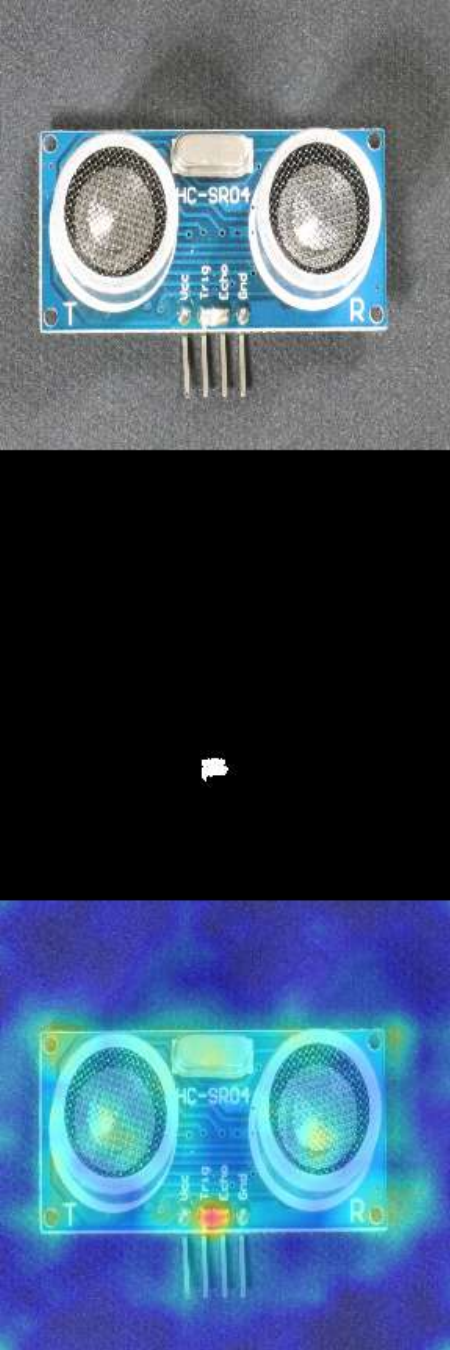}
    \includegraphics[width=0.075\textwidth]{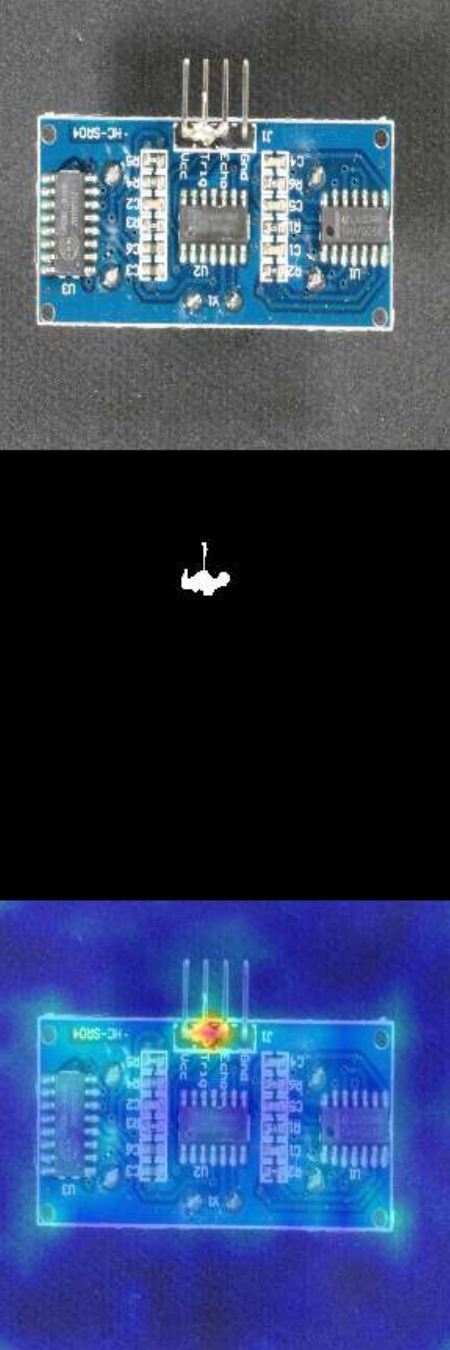}
    \includegraphics[width=0.075\textwidth]{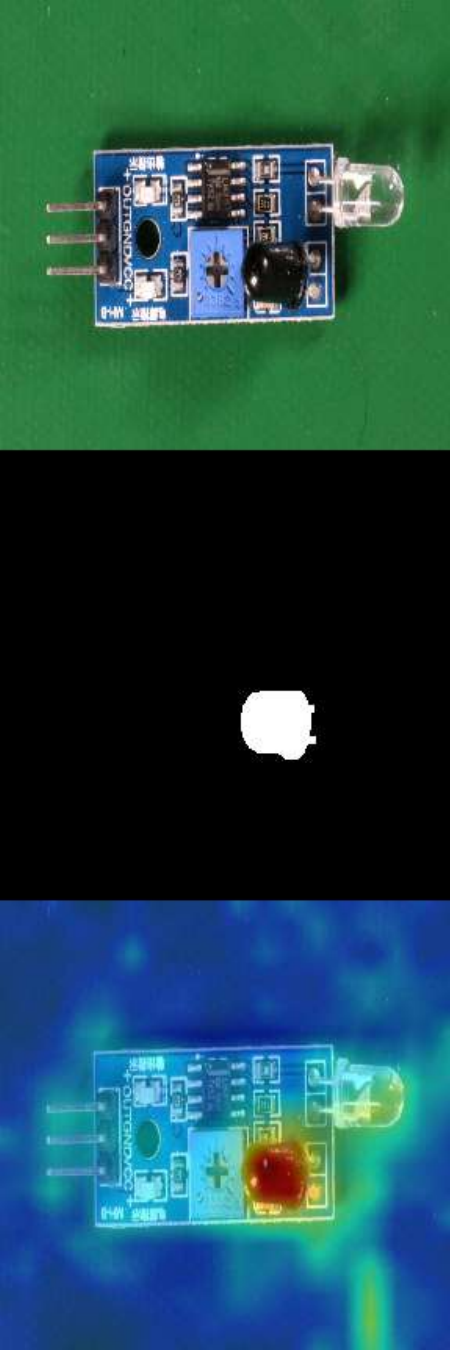}
    \includegraphics[width=0.075\textwidth]{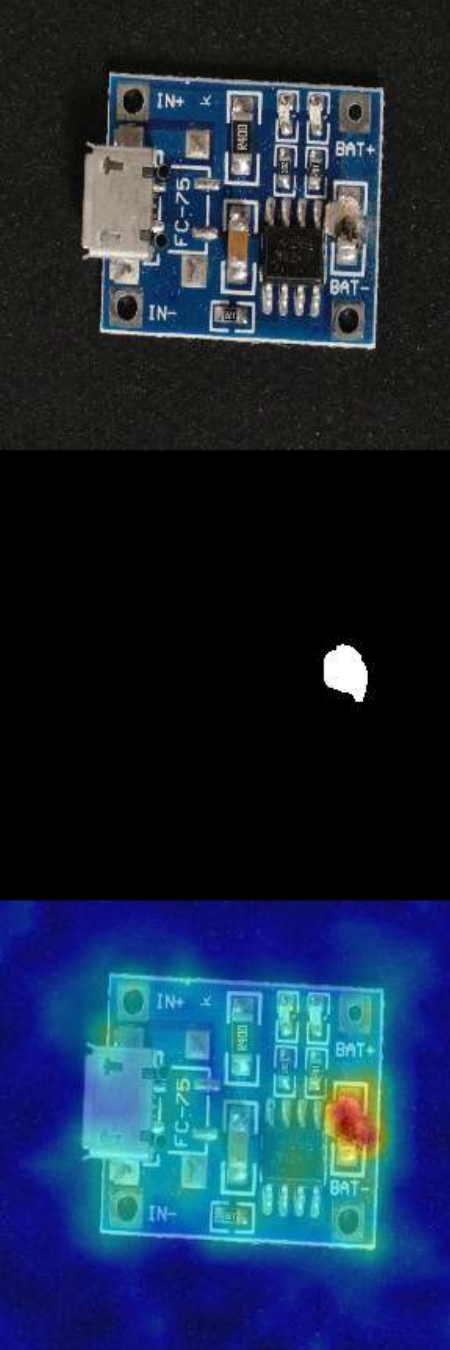}
    }
    \caption{Qualitative results, where sampled image, ground truth, and anomaly map are shown for each class in VisA.}
    \label{visa_figure}
    \vspace{-0.8em}
    
\end{figure*}

To further demonstrate the effectiveness of our proposed method, we visualize representative samples for anomaly localization of MVTec-AD using MINT-AD in Figure \ref{mvtec_figure}.
And some qualitative results of ViSA are shown in Figure~\ref{visa_figure}. 
The model is trained in the unified dataset. 

\subsection{Comparison of Unified and Separate Setting}
\label{compare-seperate}

\begin{figure}[hbt]
    \centering
    \includegraphics[width=\linewidth]{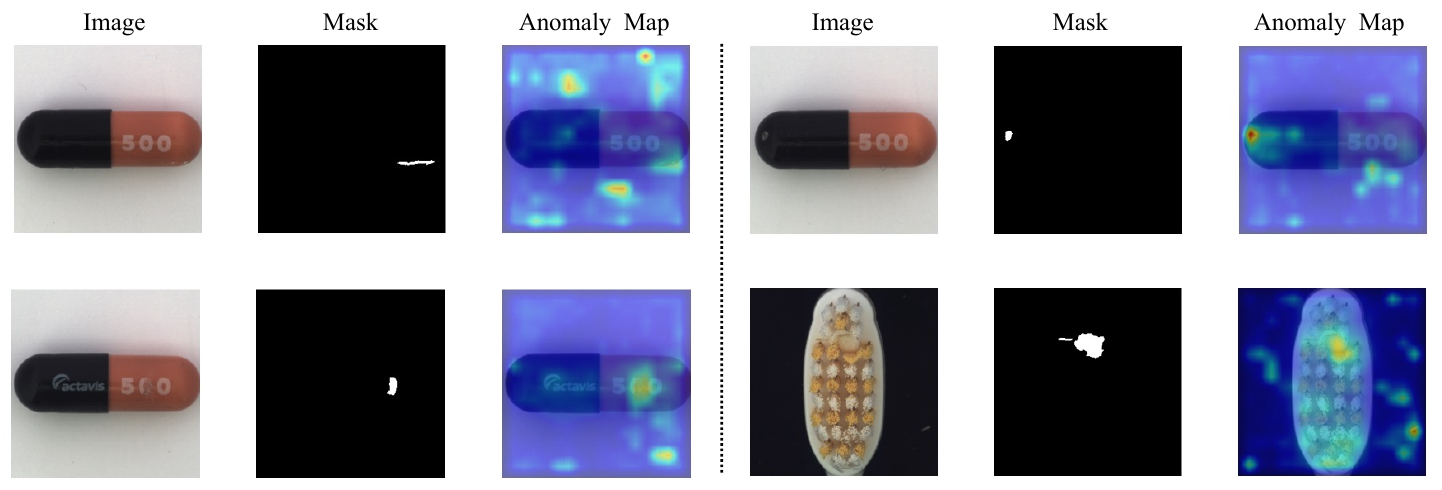}
    \caption{
    Bad cases of MINT-AD(separate setting). In the ``Capsules'' category, MINT-AD trained individually is susceptible to misjudgment due to the anomalies in the background, which is the main reason for its lower performance in this category. In other categories, such as ``Toothbrush'', this situation occurs in a few samples.}
    \label{fig:badcases}
\end{figure}

Through qualitative analysis, we discuss that the detection errors of separately trained MINT-AD in these categories are caused by background interference. As shown in Figure~\ref{fig:badcases}, the visualized bad cases of MINT-AD, it is evident that the model detects some noise in the background of the ``Capsule'' category, which masks its anomaly detection ability to objects. Interestingly, the unified training MINT-AD is more robust to this problem. We believe this is because the unified training model utilizes a sufficient amount of data to learn the reconstruction of the background. This finding suggests that future research in anomaly detection should strive to eliminate background interference as much as possible. Additionally, it reminds us that the upper limit of unified training is not the ensemble model of separately trained models. Unified training can leverage more data simultaneously, which may further enhance the performance of anomaly detection models.

\begin{table}[ht]
\centering
\caption{Anomaly detection results with AUROC metric on CIFAR-10 under the unified case. Here, \{01234\} means samples from classes 0, 1, 2, 3, and 4 are borrowed as the normal ones.}
\label{tab-cifar10}
\resizebox{0.8\linewidth}{!}{
\begin{tabular}{c|cccccc|c}
\toprule\toprule
Normal Indices & US & FCDD  & FCDD+OE  & PANDA  & MKD  & UniAD  & MINT-AD \\
\midrule$\{01234\}$ & 0.513 & 0.550 & 0.718 & 0.666 & 0.642 & 0.844 & \textbf{0.867} \\
$\{56789\}$ & 0.513 & 0.503 & 0.737 & 0.732 & 0.693 & 0.809 & \textbf{0.833} \\
$\{02468\}$ & 0.639 & 0.592 & 0.853 & 0.771 & 0.764 & 0.930 & \textbf{0.936} \\
$\{13579\}$ & 0.568 & 0.585 & 0.850 & 0.729 & 0.787 & 0.906 & \textbf{0.926} \\
\midrule 
\textbf{Mean} & 0.559 & 0.558 & 0.789 & 0.724 & 0.721 & 0.872 & \textbf{0.891} \\
\bottomrule\bottomrule
\end{tabular}
}
    \vspace{-0.8em}

\end{table}

\subsection{Anomaly detection on CIFAR-10}
For semantic anomalies, we make experiments on CIFAR-10. The experiment adopts the settings of UniAD, which consists of four combinations. For each combination, five categories collectively serve as normal samples, while other categories are treated as anomalies. The class indices refer to the class labels. The results are shown in the table \ref{tab-cifar10}, with MINT-AD outperforming other anomaly detection methods. This demonstrates the applicability of our proposed method to global semantic anomalies as well. 

\subsection{Backbone Architectures}
In this paper, we did not place excessive emphasis on the choice of backbone network. We opted for the EfficientNet-b4 network to align with the setup of UniAD for a fair comparison. However, to assess performance differences when various networks are selected, we conducted experiments on pre-trained models. The results are presented below. Across the majority of backbones, our approach consistently outperforms UniAD, highlighting the broad applicability of MINT-AD.

\begin{table}[ht]
\setlength\tabcolsep{3pt}
\caption{Ablation study on the backbone architecture. Performances on I-AUROC are reported.}
\label{tab-architecture}
\centering
\resizebox{0.85\linewidth}{!}{
\begin{tabular}{ccccc|cccc}
\toprule\toprule 
Methods & Res-18 & Res-34 & Res-50 & Res-101  & Eff-b0 & Eff-b2 & Eff-b4 & Eff-b6 \\
\midrule
UniAD & $0.924$ & $\textbf{0.930}$ & $0.924$ & $0.922$  & $0.961$ & $0.962$ & $0.965$ & $0.961$ \\
MINT-AD & \textbf{0.931}  & 0.927  & \textbf{0.943}  &  \textbf{0.931}  & \textbf{0.983}  & \textbf{0.964} & \textbf{0.986} & \textbf{0.985}  \\
\bottomrule\bottomrule
\end{tabular}
}
    \vspace{-0.8em}
\end{table}

\subsection{INR Structure}

From a foundational perspective, INR (Implicit Neural Representation) is generally based on the perceptron network structure. However, it takes advantage of coordinates as inputs, making it more conducive to modeling the given data. In this paper, our intention is to model the distribution of normal features. Using MLP is not efficient for modeling this spatial information. Moreover, we recognize that individual features maintain a positional relationship on the feature map. Thus, using continuous positional coordinates as inputs is more beneficial for modeling. 

Moreover, our INR is designed as a dual-MLP architecture which permits the INR to model multi-class feature distribution in one model. By taking in class tokens $t_c$ and a single positional encoding $z$, the INR network maps them to the feature dimensions, $ \phi_{\text{query}} = \text{INR}(t_c, z) $. The INR consists of both a Synthesis Network and a Modulation Network. The Synthesis Network defines a continuous function that maps positional encodings to features. Notation output of each layer as $ h_1, ...,h_K $, the output of each layer serves as the input for the next layer.
\begin{equation}
    h_0 = z, h_i=\alpha_i \odot \sin \left(w_i h_{i-1}+b_i\right),
\end{equation}
where $ w $ and $ b $ are weights and biases, $\odot$ is element-wise multiplied, while $ \alpha_i $ corresponds to the output of each layer in the Modulation Network. For the layers of the Modulation Network,
\begin{equation}
\alpha_0 = z, \alpha_i = \text{RELU}(w_i'[\alpha_{i-1}, t_z]^T +b_i').
\end{equation}
The two networks employ \textit{sin} and ReLU as their activation functions, respectively. Mapping each position results in a complete query, denoted as $ \boldsymbol{f_{\text{query}}} $, which is used as an input to the reconstruction network. The distribution encoder then maps each feature to the posterior distribution parameters of a single position, $ \omega = \text{Decoder}(\phi_{\text{query}}) $.

We added an experiment to demonstrate this. In the experiment, we replaced the INR with an MLP to get the class-aware query. $\boldsymbol{f_{\text{query}}}=Repeat(MLP(t_c))$, MLP map the class token to feature dimension and repeat it to get the full query map. We also remove the corresponding prior loss. The results in Table~\ref{tab:mlp} show that the performance of using an MLP-based query is reduced compared to our INR at the image level. At the same time, the parameter count of MLP is quite large, which imposes an additional computational burden on the system.

\begin{table}[ht]
\centering
\caption{Different class-aware query structures.}
\resizebox{0.65\linewidth}{!}{
    \begin{tabular}{lccc}\\\toprule\toprule
    Modules & I-AUROC & P-AUPR & Params of module (M) \\\midrule
    MLP-Query & 0.980 & \textbf{0.498} & 383.5 \\ 
    INR-Query  & \textbf{0.986}  & \textbf{0.498} & \textbf{0.602} \\  \bottomrule\bottomrule
    \end{tabular}
   \label{tab:mlp}
   }
    \vspace{-0.8em}
\end{table}

\subsection{Anomaly Localization on MVTec AD}
Table~\ref{tab-mint-mvtec-locliaztion} displays the performance of MINT-AD in anomaly localization on the MVTec AD dataset. It is worth noting that the actual resolution of the anomaly map output by MINT-AD is 14x14(interpolating it to the target size). Under this premise, MINT-AD's localization performance surpasses most methods that do not have additional optimization for localization. However, we also believe that improving the output resolution can further enhance performance. Some existing studies have explored various approaches, such as increasing image resolution and network size. Additionally, some research has achieved better performance by incorporating an additional segmentation network to output localization results~\cite{7,li2023efficient,Lei_2023_CVPR}. This paper primarily focuses on anomaly classification performance in multi-class anomaly detection, so these techniques are areas for future exploration.

\begin{table}[htb]
    \vspace{-0.8em}
\centering
\setlength\tabcolsep{3pt}
\caption{Performance Comparison of Different Methods in Multi-class Anomaly Localization with MVTec AD. (Metrics: P-AUPR)}
\resizebox{\linewidth}{!}{
\begin{tabular}{ll|cccccc}
\toprule\toprule
\multicolumn{2}{c|}{Categories} & CFA~\cite{lee2022cfa} & UTRAD~\cite{chen2022utrad} & CFLOW~\cite{gudovskiy2022cflow} & SimpleNet~\cite{liu2023simplenet} & UniAD~\cite{you2022unified} & MINT-AD(ours) \\ \midrule
\multirow{10}{*}{\rotatebox{90}{Objects}} 
        & Bottle & 0.595 & 0.375 & 0.099 & 0.371 & 0.655 & \textbf{0.665} \\ 
        ~ & Cable & 0.469 & 0.654 & 0.048 & 0.044 & \textbf{0.479} & 0.405 \\ 
        ~ & Capsule & 0.394 & 0.208 & 0.052 & 0.085 & 0.438 & \textbf{0.475} \\ 
        ~ & Hazelnut & 0.518 & 0.424 & 0.038 & 0.171 & 0.554 & \textbf{0.572} \\ 
        ~ & Metal nut & \textbf{0.877} & 0.543 & 0.162 & 0.374 & 0.549 & 0.615 \\ 
        ~ & Pill & \textbf{0.759} & 0.492 & 0.066 & 0.286 & 0.445 & 0.527 \\ 
        ~ & Screw & 0.160 & 0.073 & 0.011 & 0.011 & 0.296 & \textbf{0.512} \\ 
        ~ & Toothbrush & \textbf{0.456} & 0.202 & 0.023 & 0.050 & 0.373 & 0.380 \\ 
        ~ & Transistor & \textbf{0.687} & 0.580 & 0.106 & 0.131 & 0.664 & 0.652 \\ 
        ~ & Zipper & 0.458 & 0.257 & 0.050 & 0.184 & 0.403 & \textbf{0.583} \\ \midrule
\multirow{5}{*}{\rotatebox{90}{Textures}}  
        & Carpet & 0.467 & 0.235 & 0.166 & 0.309 & 0.478 & \textbf{0.556} \\ 
        ~ & Grid & 0.215 & 0.117 & 0.205 & 0.003 & 0.226 & \textbf{0.340} \\ 
        ~ & Leather & \textbf{0.381} & 0.115 & 0.086 & 0.181 & 0.331 & 0.369 \\ 
        ~ & Tile & \textbf{0.442} & 0.366 & 0.172 & 0.433 & 0.415 & 0.417 \\ 
        ~ & Wood & 0.381 & 0.199 & 0.121 & 0.258 & 0.372 & \textbf{0.400} \\ \midrule
        ~ & \textbf{Mean} & 0.483 & 0.302 & 0.094 & 0.193 & 0.445 & \textbf{0.498} \\  \bottomrule\bottomrule
    \end{tabular}
}
\label{tab-mint-mvtec-locliaztion}
    \vspace{-0.8em}
\end{table}

\section{Method Analysis}

\subsection{Correctness of Prior Loss}
Using the Lemma 1 in IGD~\cite{104}, the maximization of the constrained
$\mathcal{L}_{E L B O}(q, \theta_a)$ 
that makes 
$\mathbb{E}_{q(\omega)}[\log \mathcal{P}\left( \boldsymbol{f}_c \mid t_{c}, \theta \right)] \geq \mathbb{E}_{q(\omega)}[\log \mathcal{P}\left( \boldsymbol{f}_c \mid t_{c}, \theta_{old} \right)]$ 
can be estimated by EM optimization. The E-step is to estimate $\omega$ of the prior distribution $q(\omega)$ using statistics, and the M-step is to learn an effective normality description. The M-step optimizes $\theta_p$ by gradient descent, and the E-step can also be approximated with $\theta_a$ by gradient descent~\cite{neal1998view}. So $\log \mathcal{P}\left( \boldsymbol{f}_c \mid t_{c}, \theta \right) - \log \mathcal{P}\left( \boldsymbol{f}_c \mid t_{c}, \theta_{old} \right)$ is lower bounded by $(\mathbb{E}_{q(\omega)}[\log \mathcal{P}\left( \boldsymbol{f}_c \mid t_{c}, \theta \right)] - \mathbb{E}_{q(\omega)}[\log \mathcal{P}\left( \boldsymbol{f}_c \mid t_{c}, \theta_{old} \right)]) \geq 0$.

\subsection{Architecture Details}
In Table~\ref{tab-backbone}, we present the architecture of the backbone network, which processes a batch of images and produces a corresponding batch of feature embeddings with the dimensions of 14×14×324. Table~\ref{tab-prompt-mapper} details the architecture of the prompt mapper, a classifier that generates a class token at the penultimate layer. Table~\ref{tab-INR} outlines the parameters of the Implicit Neural Representation (INR), wherein the synthesis network accepts position embeddings as input, and the modulation network utilizes concatenated embeddings of position and class token. Table~\ref{tab-distribution-decoder} describes the architecture of the distribution decoder. Finally, Table~\ref{tab-reconstruction-transformer} illustrates the architecture of the reconstruction transformer.

\begin{table}[ht]
\setlength\tabcolsep{3pt}
\centering
\caption{Architecture of feature extractor. This backbone follows EfficinetNet-B4, albeit with only the initial 6 stages utilized. Each row describes a stage $i$ with $\hat{L}_i$ layers, with input resolution $\hat{H}_i \times \hat{W}_i$ and output channels $\hat{C}_i$.}
\resizebox{0.8\linewidth}{!}{

\begin{tabular}{c|c|c|c|c}
\toprule
Stage $i$ & Operator $\hat{F}_i$ & Resolution $\hat{H}_i \times \hat{W}_i$ & \#Channels $\hat{C}_i$ & \#Layers $\hat{L}_i$ \\ \midrule
1         & Conv3$\times$3                                   & 224 $\times$ 224                                 & 32                     & 1                   \\ 
2         & MBConv1, k3$\times$3                             & 112 $\times$ 112                                 & 16                     & 1                   \\ 
3         & MBConv6, k3$\times$3                             & 112 $\times$ 112                                 & 24                     & 2                   \\ 
4         & MBConv6, k5$\times$5                             & 56 $\times$ 56                                   & 40                     & 2                   \\ 
5         & MBConv6, k3$\times$3                             & 28 $\times$ 28                                   & 80                     & 3                   \\ 
6         & MBConv6, k5$\times$5                             & 14 $\times$ 14                                   & 112                    & 3                   \\ \bottomrule
\end{tabular}}
\label{tab-backbone}
    \vspace{-0.8em}
\end{table}

\begin{table}[htb]
    \vspace{-0.8em}

\setlength\tabcolsep{3pt}
\centering
\caption{Architecture of prompt mapper. Notice that we use the output of the channel encoder as the prompt, but not of the head. }
\resizebox{0.9\linewidth}{!}{
\begin{tabular}{c|c|c|c}
\toprule
Stage & Operator & \#Parameters  & Activation \\ \midrule
Operation & Reshape & B $\times$ C $\times$ (H $\times$ W)->B $\times$ C $\times$ P & - \\
Class Encoder & Linear & 324 $\times$ 256  &  ReLU \\ 
              & Linear & 256 $\times$ 256  &  ReLU \\ 
              & Linear & 256 $\times$ 32  &  Softma$\times$ \\ \hline
Operation & Reshape & B $\times$ (C $\times$ P$'$) -> B $\times$ P$^*$ & - \\
Channel Encoder & Linear & 8192 $\times$ 256  &  LeakyReLU \\ 
              & Linear & 256 $\times$ 256  &  LeakyReLU \\ 
              & Linear & 256 $\times$ 32  &  Softmax \\ \hline
Classification Head    & Linear & 32 $\times$ 15 (for MVTec AD)  &  - \\ 
\bottomrule
\end{tabular}
\label{tab-prompt-mapper}}
    \vspace{-0.8em}
\end{table}

\begin{table}[ht]
\setlength\tabcolsep{3pt}
\centering
\caption{Architecture of INR. Each stratum's output within the Synthesis Network undergoes multiplication with the output of the Modulation Network before proceeding to the subsequent stratum. }
\begin{tabular}{c|c|c|c}
\toprule
Stage & Operator & \#Parameters  & Activation \\ \midrule
Synthesis Network & Linear & 3 $\times$ 256  &  Sine \\ 
              & Linear $\times$ 4 & 256 $\times$ 256  &  Sine \\ \hline
Modulation Network & Linear & (32 + 3) $\times$ 256  &  ReLU \\ 
              & Linear $\times$ 4 & 256 $\times$ 256  &  ReLU \\ 
\bottomrule
\end{tabular}
\label{tab-INR}
    \vspace{-0.8em}
\end{table}

\begin{table}[ht]
\setlength\tabcolsep{3pt}
\centering
\caption{Architecture of distribution decoder. The output of the decoder is split to {mean, variance} of the distribution}
\begin{tabular}{c|c|c|c}
\toprule
Stage & Operator & \#Parameters  & Activation \\ \midrule
Distribution Decoder & Linear & 256 $\times$ 256  &  LeakyReLU \\ 
              & Linear  & 256 $\times$ 256  &  LeakyReLU \\ 
              & Linear  & 256 $\times$ 512  &  - \\ 
\bottomrule
\end{tabular}
\label{tab-distribution-decoder}
    \vspace{-0.8em}
\end{table}

\begin{table}[ht]
\setlength\tabcolsep{3pt}
\centering
\caption{Architecture of reconstruction transformer.}
\begin{tabular}{c|c|c}
\toprule
Stage & Operator & \#Parameters \\ \midrule
EncoderLayer $\times$ 4 &  SelfAttention & 256 $\times$256 \\
                & Liner1  & 256 $\times$ 1024   \\ 
              & Dropout1  & 0.1 \\ 
              & LayerNorm1 & 256 \\
              & Linear2  & 1024 $\times$ 256 \\ 
              & Dropout2  & 0.1 \\ 
              & LayerNorm2 & 256 \\\hline
DecoderLayer $\times$ 4 &  SelfAttention & 256 $\times$256 \\
              & Dropout1  & 0.1 \\ 
              & LayerNorm1 & 256 \\
              &  MultiheadAttention & 256 $\times$256 \\
              & Dropout2  & 0.1 \\ 
              & LayerNorm2 & 256 \\
              & Liner1  & 256 $\times$ 1024   \\ 
              & ReLU & \\
              & Dropout2  & 0.1 \\ 
              & Linear2  & 1024 $\times$ 256 \\ 
              & Dropout3  & 0.1 \\ 
              & LayerNorm3 & 256 \\
\bottomrule
\end{tabular}
\label{tab-reconstruction-transformer}
    \vspace{-0.8em}
\end{table}

\clearpage

\end{document}